\pdfoutput=1

\documentclass[11pt]{article}

\usepackage[preprint]{acl}

\usepackage{times}
\usepackage{latexsym}

\usepackage[T1]{fontenc}

\usepackage[utf8]{inputenc}

\usepackage{microtype}

\usepackage{inconsolata}

\usepackage{graphicx}
\usepackage{subcaption}
\usepackage{tcolorbox}
\usepackage{booktabs}
\usepackage{amsmath}
\graphicspath{{pdfs/}}

\title{From Parameters to Performance: A Data-Driven Study \\ on LLM Structure and Development}

\author{
    Suqing Wang$^{2}$\footnotemark[2], 
    Zuchao Li$^{1}$\footnotemark[1]\footnotemark[2],
    Luohe Shi$^{2}$, 
    Bo Du$^{2}$, 
    Hai Zhao$^{3}$, \\ 
    \textbf{Yun Li}$^{4}$ \textbf{and} \textbf{Qianren Wang}$^{4}$\\
    {$^{1}$School of Artificial Intelligence, Wuhan University} \\
    {$^{2}$School of Computer Science, Wuhan University} \\
    {$^{3}$School of Computer Science, Shanghai Jiao Tong University} \\
    {$^{4}$Cognitive AI Lab, Shanghai Huawei Technologies, China} \\
    {\tt \{wangsuqing,zcli-charlie,shiluohe,dubo\}@whu.edu.cn} \\
    {\tt zhaohai@cs.sjtu.edu.cn,lychina@139.com,wangqr2019@qq.com}
}

\begin{document}
\maketitle
\renewcommand{\thefootnote}{\fnsymbol{footnote}} 
\footnotetext[1]{Corresponding author.} 
\footnotetext[2]{Equal contribution. }
\renewcommand{\thefootnote}{}
\begin{abstract}
Large language models (LLMs) have achieved remarkable success across various domains, driving significant technological advancements and innovations. Despite the rapid growth in model scale and capability, systematic, data-driven research on how structural configurations affect performance remains scarce. To address this gap, we present a large-scale dataset encompassing diverse open-source LLM structures and their performance across multiple benchmarks. Leveraging this dataset, we conduct a systematic, data mining-driven analysis to validate and quantify the relationship between structural configurations and performance. Our study begins with a review of the historical development of LLMs and an exploration of potential future trends. We then analyze how various structural choices impact performance across benchmarks and further corroborate our findings using mechanistic interpretability techniques. By providing data-driven insights into LLM optimization, our work aims to guide the targeted development and application of future models. We will release our dataset at \url{https://huggingface.co/datasets/DX0369/LLM-Structure-Performance-Dataset}.
\end{abstract}

\begin{table*}
  \centering
  \setlength{\tabcolsep}{4pt}
  \begin{tabular}{l|lllllllll}
    \hline
    \textbf{Column} & \textbf{Mean} & \textbf{Mode} & \textbf{Q1} & \textbf{Q2} & \textbf{Q3} & \textbf{Max} & \textbf{Skewness} & \textbf{Kurtosis} & \textbf{Miss Rate} \\
    \hline
    size & 8 & 8 & 1 & 7 & 8 & 1018 & 12 & 357 & 18\% \\
    d\_model & 3284 & 4096 & 2048 & 4096 & 4096 & 50257 & 0 & 5 & 5\% \\
    d\_ffn & 12767 & 14336 & 9216 & 14336 & 14336 & 13100072 & 343 & 120913 & 21\% \\
    heads & 28 & 32 & 16 & 32 & 32 & 5000 & 124 & 32475 & 5\% \\
    layers & 30 & 32 & 24 & 32 & 32 & 8928 & 187 & 49768 & 5\% \\
    kv\_heads & 15 & 8 & 8 & 8 & 32 & 160 & 1 & 1 & 29\% \\
    vocab\_size & 76579 & 32000 & 32000 & 50257 & 128256 & 5025700 & 4 & 272 & 4\% \\
    pos & 30913 & 4096 & 2048 & 4096 & 32768 & 104857600 & 271 & 85268 & 7\% \\
    downloads & 1827 & 10 & 10 & 14 & 21 & 24279491 & 171 & 36681 & 5\% \\
    likes & 2 & 0 & 0 & 0 & 0 & 5927 & 61 & 5392 & 5\% \\
    \hline
  \end{tabular}
    \caption{
      Statistical summarization of our proposed dataset, includes various statistics for model structure attributes, including \textbf{Mean}, \textbf{Mode}, \textbf{Q1} (first quartile), \textbf{Q2} (the middle value of the dataset), \textbf{Q3} (third quartile), \textbf{Skewness} (measure of asymmetry in the distribution), \textbf{Kurtosis} (measure of the "tailedness" of the distribution), and \textbf{Miss Rate} (percentage of missing values in the dataset). 
    }
  \label{model architectural statistics}
\end{table*}

\section{Introduction} 
Large language models (LLMs) have revolutionized a wide range of domains, including natural language understanding and generation~\citep{radford2019language, li2025dialogue}, as well as multimodal applications~\citep{achiam2023gpt, yang2024emollm, yang2024synergy, he2024multi}, driving significant advancements in both technology and real-world applications. Models such as GPT-3~\citep{brown2020language}, Qwen~\citep{bai2023qwen}, and LLaMA~\citep{touvron2023llama} have demonstrated outstanding performance by leveraging scaling laws~\citep{kaplan2020scaling}, which link improvements in model performance with increases in model size, training data, and computational resources. These models have set new benchmarks across various fields. However, despite the remarkable progress in scaling up these models, a systematic exploration of the relationship between structural configurations and task-specific performance remains lacking.

As LLMs become increasingly complex and resource-intensive, deploying these models in real-world applications presents significant challenges in terms of cost and energy consumption~\citep{zhao2023survey, kaddour2023challenges}. In response, the field is actively exploring efficiency optimization techniques, with prominent examples including KV-Cache reduction~\citep{tang2025spindlekv, zhao2025iam, shi2024keep} and various model lightweighting methods~\citep{ma2025model, yang2025faster}. While structural configurations are known to influence model performance~\citep{yang2024unveiling, dong2023abilities}, their effects across different tasks and application domains have not been comprehensively analyzed, with discussions often limited to qualitative hypotheses or small-scale experiments. The growing complexity of LLMs necessitates a deeper exploration of the trade-offs between various structural designs, computational resources, and model performance, calling for quantitative validation of previous hypotheses and explorations.

To address these challenges, we present a large-scale dataset encompassing various open-source LLMs' structural configurations and their performance across multiple benchmarks, providing a foundation for data-driven insights into the relationship between model structure and performance. This paper reviews the historical development of LLMs and explores how structural configurations impact LLMs' performance. Additionally, we employ mechanistic interpretability techniques to investigate the mechanism of models across diverse benchmarks, further validating the phenomena uncovered in the dataset. Through this analysis, by providing the first large-scale, quantitative validation for previous hypotheses and transforming qualitative conjectures into measurable conclusions, we offer valuable data-driven insights for optimizing LLMs design, contributing to the development of models that are not only powerful and scalable but also efficient and adaptable to diverse applications. We will release our code at \url{https://github.com/DX0369/llm-structure-performance}.

\begin{figure}[t]
    \centering
    \includegraphics[width=0.47\textwidth]{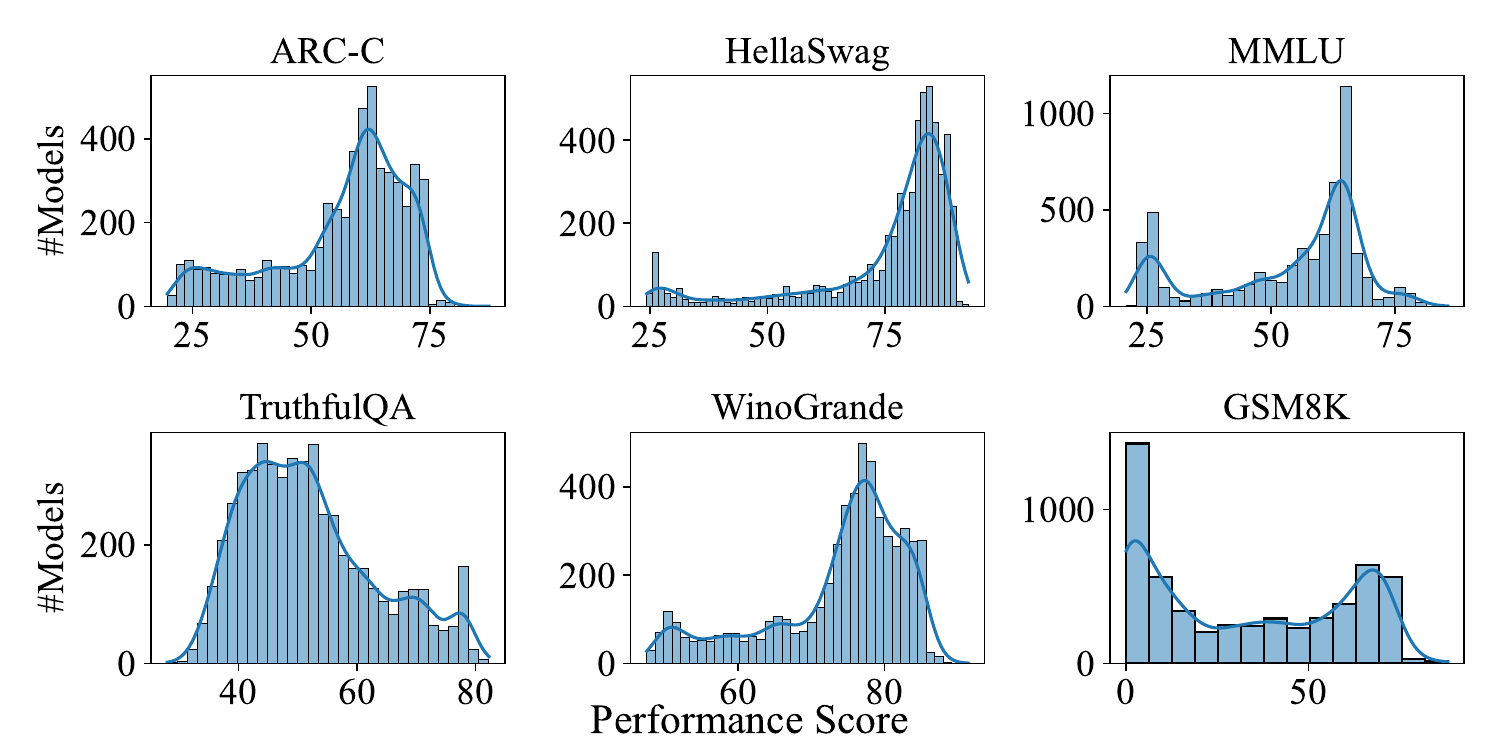}
    \caption{The performance score distributions of open-source LLMs across six benchmarks in our LLMs Structure and Performance Dataset, which illustrate overall performance trends. The x-axis represents performance scores, while the y-axis indicates the number of models achieving each score.}
    \label{fig:performance_distribution}
\end{figure}

Our key contributions are summarized as follows:
\begin{itemize}
    \item \textbf{Large-Scale Open-Source LLMs Structure and Performance Dataset:} We introduce a large-scale dataset containing a variety of open-source LLMs' structural configurations and their performance on multiple benchmarks, offering a foundation for data-driven insights into the relationship between model structure and performance.
    \item \textbf{Study on the Impact of Structure on Performance:} We provide the first large-scale, quantitative validation of how structural configurations influence LLM performance, offering robust empirical evidence for the roles of key parameters like layer depth.
    \item \textbf{Mechanistic Interpretability Analysis and Validation:} We employ layer-pruning and gradient analysis techniques to validate the findings regarding the impact of layer depth on performance across different benchmarks, as mined from the LLMs Structure and Performance Dataset.
\end{itemize}

\section{LLMs Structure and Performance Dataset} 
Our dataset is sourced from the Hugging Face model database and the Open LLM Leaderboard. Model structure details are retrieved from structured configuration files of models available on Hugging Face.

For model structural configuration, our dataset primarily includes \texttt{size} (model size), \texttt{d\_model} (hidden dimension), \texttt{d\_ffn} (FFN intermediate size), \texttt{heads} (number of attention heads), \texttt{layers} (layer depth), \texttt{date} (publication date), and, as an additional feature, \texttt{likes} (the number of user likes on Hugging Face model pages).

For model performance, we extract evaluation results from the Open LLM Leaderboard v1, which provides performance metrics for open-source LLMs across six widely used benchmarks
: ARC-Challenge~\citep{clark2018think}, HellaSwag~\citep{zellers2019hellaswag}, MMLU~\citep{hendrycks2020measuring}, TruthfulQA~\citep{lin2021truthfulqa}, WinoGrande~\citep{sakaguchi2021winogrande}, and GSM8K~\citep{cobbe2021training}.

The collected data is cleaned and manually verified. Models that are no longer available are removed, and missing data is supplemented through technical reports or source code, ensuring accuracy. Additionally, potential errors are cross-checked during this process. We identify and label the models that are Mixture of Experts (MoE) or multimodal models. The dataset consists of approximately 160,000 model configuration entries, among which about 6,000 entries contain performance metrics. These performance records focus on representative and widely adopted checkpoints rather than numerous derivative variants, making them sufficient to support reliable and generalizable analyses. The statistical properties of the model structure are summarized in \autoref{model architectural statistics}, while the performance score distribution is shown in Figure~\ref{fig:performance_distribution}. The details of the dataset can be found in Appendix~\ref{app:Details}.

\section{Trends Uncovered from Data Analysis} 
\begin{figure}[t]
    \centering
    \includegraphics[width=0.47\textwidth]{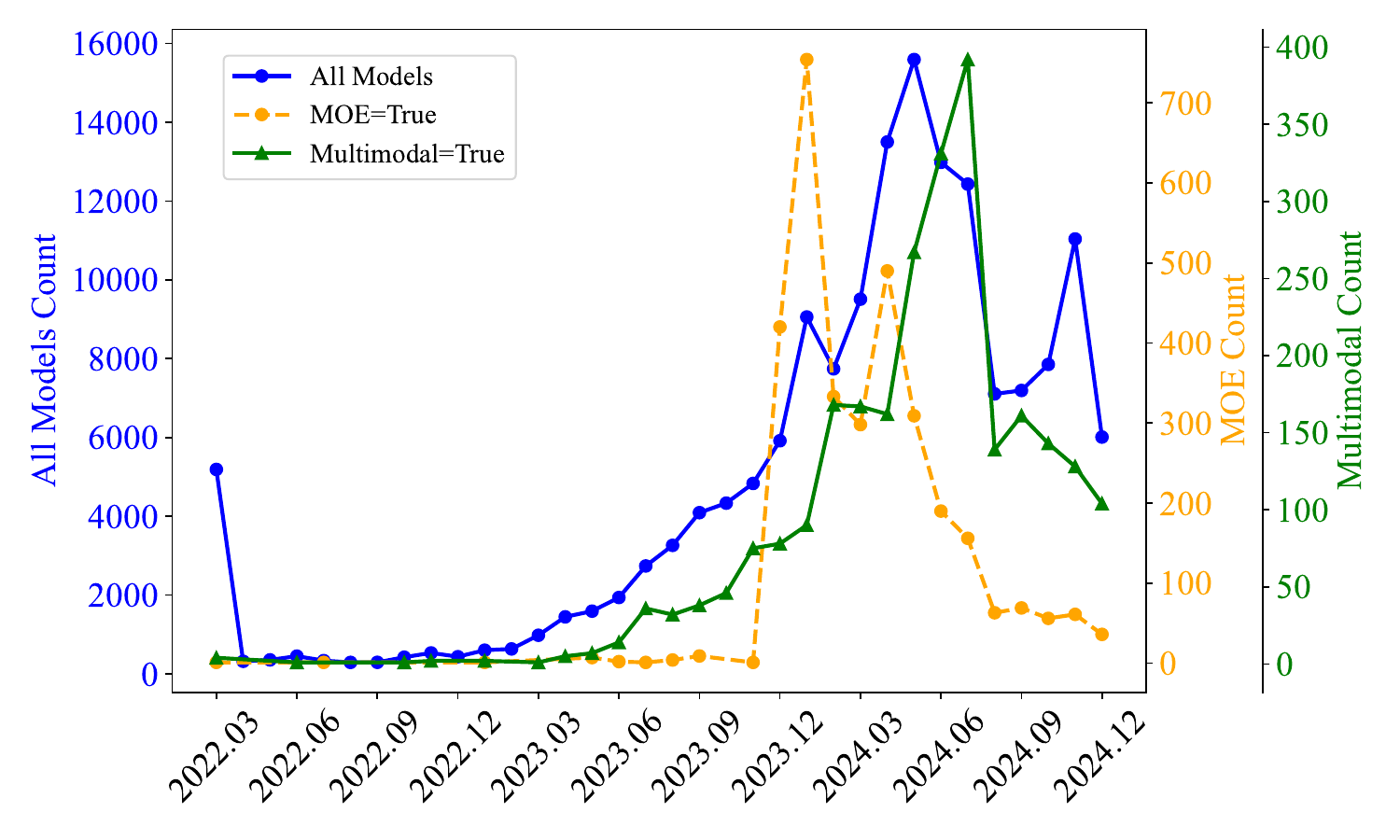} 
    \caption{Monthly count distribution of new open-source LLMs: MoE, multimodal, and all models over time.}
    \label{fig:model_count_by_category}
\end{figure}

\textbf{The growth rate of MoE models has slowed, while multimodal models continue to be widely popular.}
We analyze the monthly variations in the number of LLMs across different categories, as shown in Figure~\ref{fig:model_count_by_category}. Since the release of ChatGPT in November 2022, the number of LLMs has surged rapidly. The trend in multimodal LLMs mirrors that of overall LLMs. In contrast, models based on the MoE architecture saw a sharp increase after the release of Mixtral 8x7B~\citep{jiang2024mixtral} in December 2023. However, its growth rate slowed after six months. Although Deepseek and Qwen have open-sourced smaller models better suited for private deployment~\citep{dai2024deepseekmoeultimateexpertspecialization, yang2024qwen2}, MoE models not only require more resources than dense models, but their load balancing requirements also introduce greater fine-tuning challenges, such as instability~\citep{dai2022stablemoestableroutingstrategy}.

\begin{figure*}[t]
    \centering
    \begin{subfigure}{0.46\textwidth}
        \centering
        \includegraphics[width=\textwidth]{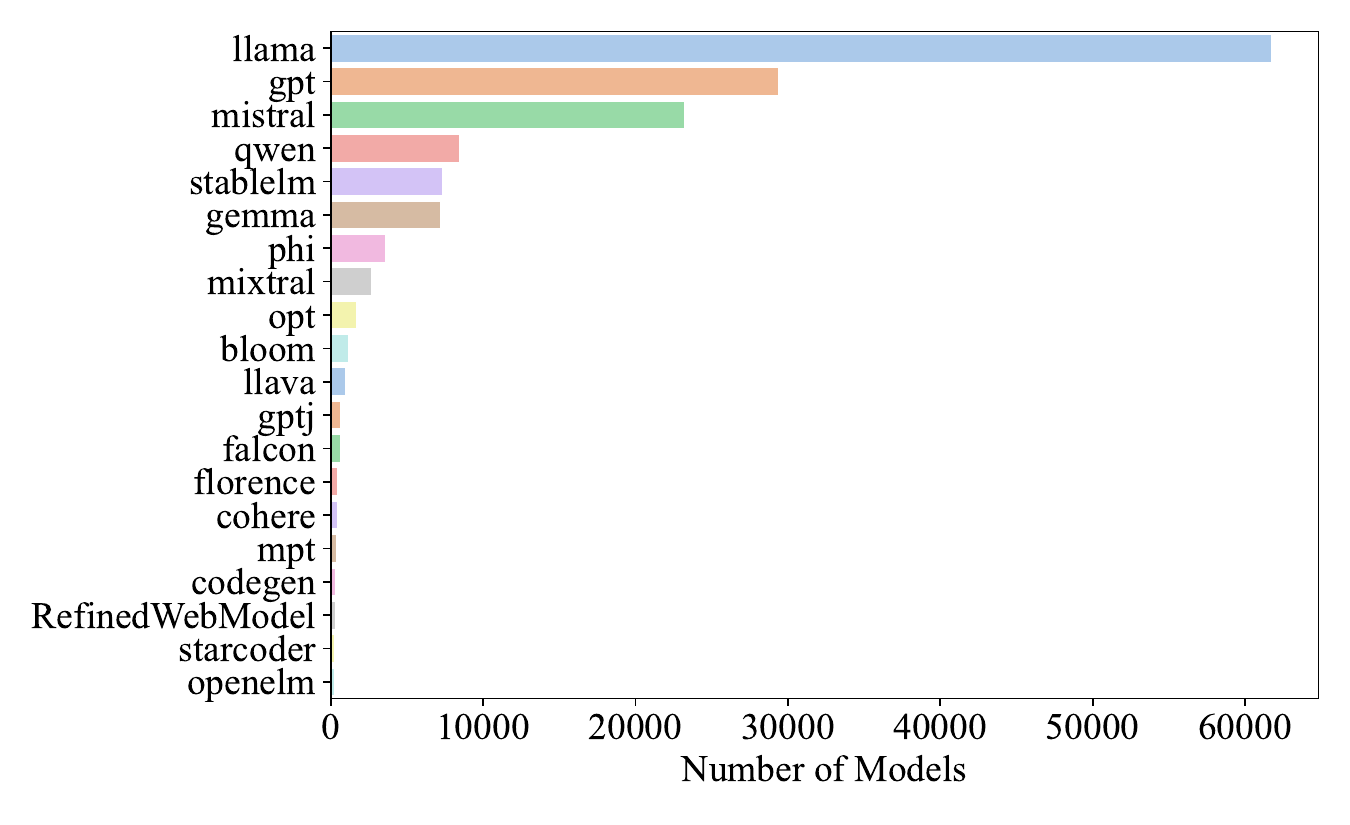}
        \caption{}
        \label{fig:order_by_frequency}
    \end{subfigure}
    \begin{subfigure}{0.46\textwidth}
        \centering
        \includegraphics[width=\textwidth]{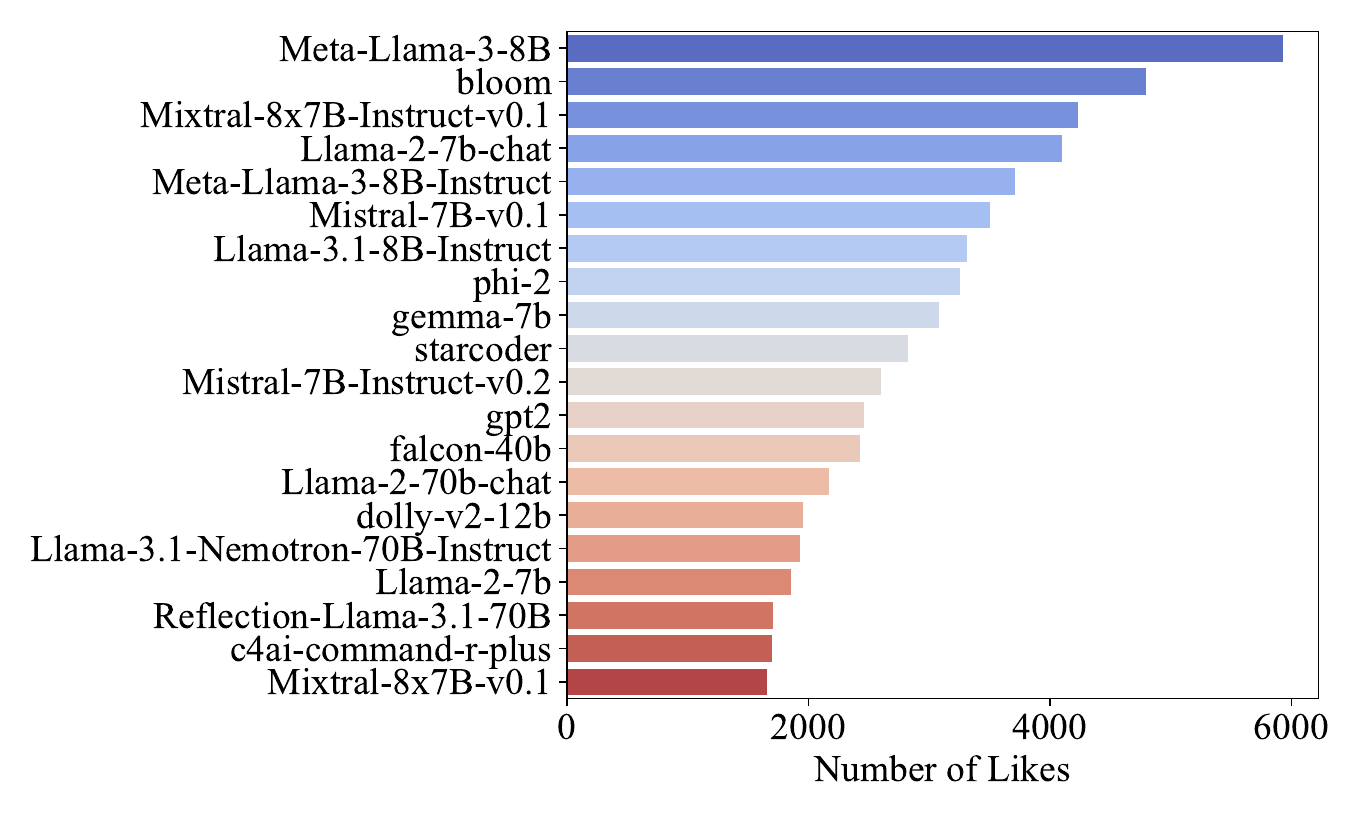}
        \caption{}
        \label{fig:order_by_likes}
    \end{subfigure}
    \caption{(a) Top 20 types of open-source LLMs sorted by model count. (b) Top 20 open-source LLMs sorted by the number of likes.}
\end{figure*}

\begin{figure}[t]
    \centering
    \includegraphics[width=0.47\textwidth]{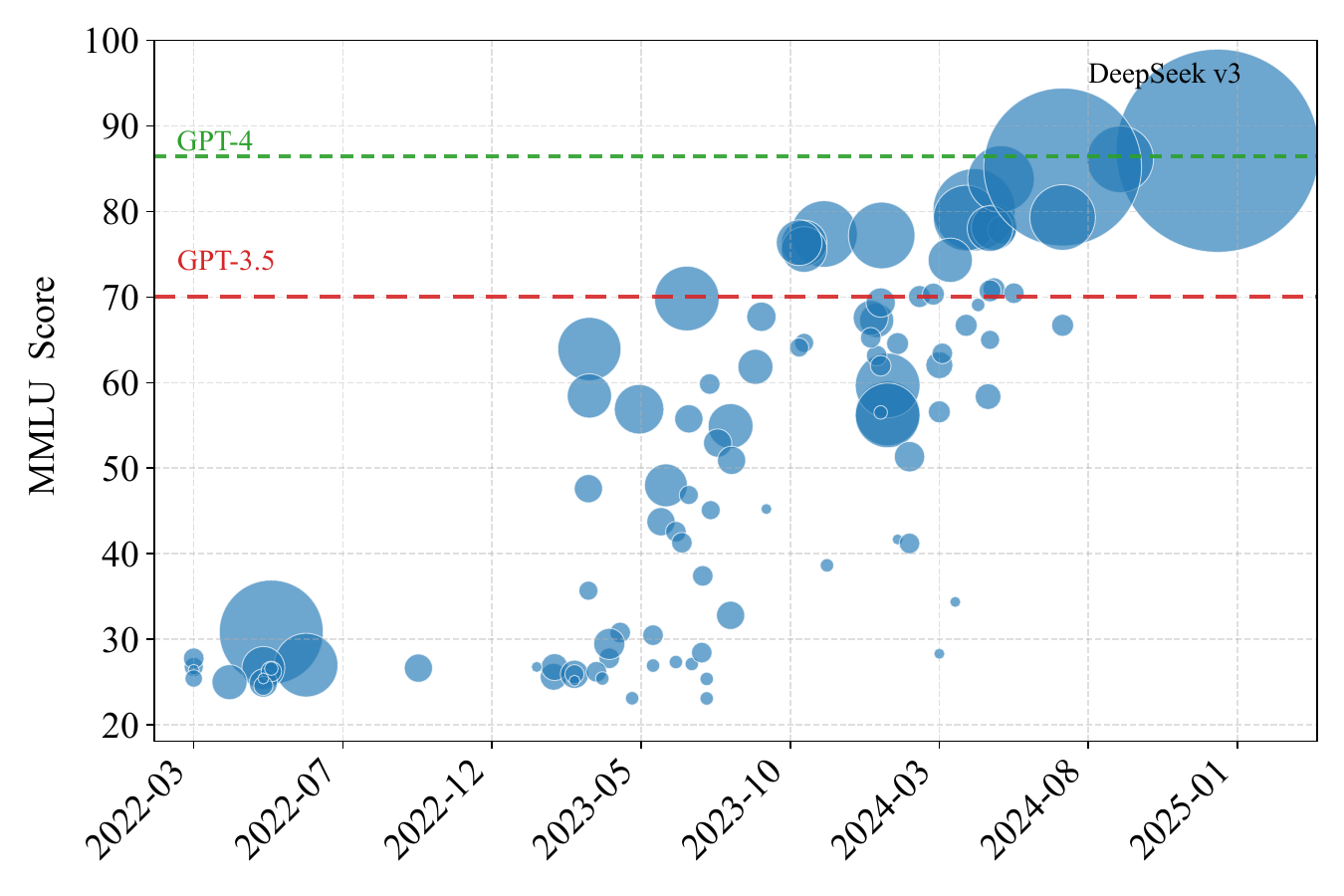}
    \caption{The performance evolution of major open-source pre-trained models in the MMLU over time, where the size of the data points reflects the model scale.}
    \label{fig:all_average_performance_pre-trained_models_over_time}
\end{figure}

\begin{figure*}[t]
    \centering
    \includegraphics[width=0.75\textwidth]{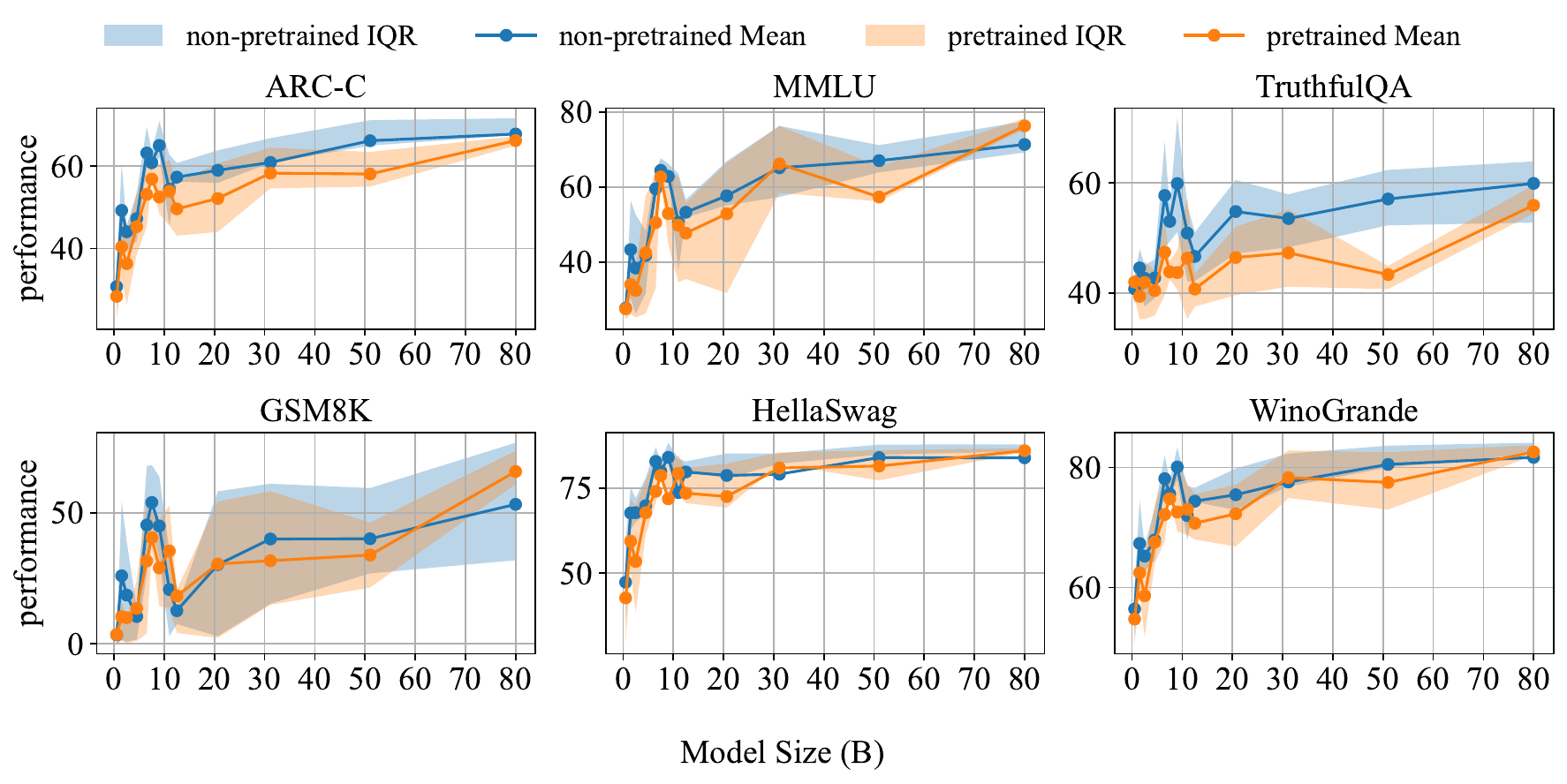}
    \caption{Performance of different datasets across different model size and training  strategies, with equal-frequency binning and interquartile range (IQR) shading to capture performance variation.}
    \label{fig:performance_size_train}
\end{figure*}

\begin{figure}[t]
    \centering
    \includegraphics[width=0.47\textwidth]{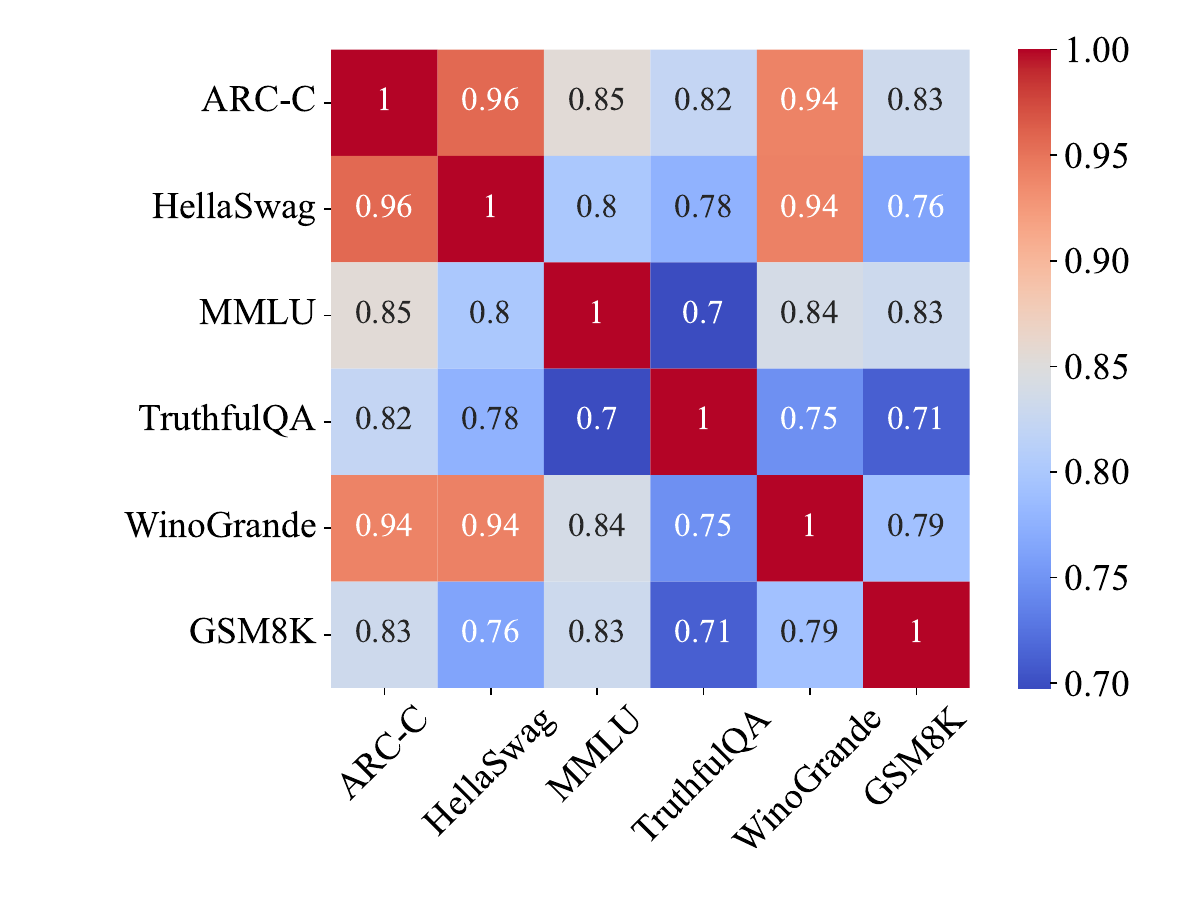}
    \caption{Spearman rank correlation coefficients matrix of performance across different benchmarks.}
    \label{fig:performance_spearman_rank_correlation}
\end{figure}

\textbf{LLaMA are the most popular base model.}
Analyzing open-source LLM model types, such as {\tt NameForCausalLM}, provides insights into the base models used for fine-tuning, as shown in Figure~\ref{fig:order_by_frequency}. Here we count the number of derivative models within each model family, which reflects the extent to which a base model has been adopted and diversified in the community. LLaMA is the most widely adopted base model, followed by the GPT series. Mistral, originating from Europe, ranks third.

\textbf{7B-scale and 70B-scale models are the most popular.}
Figure~\ref{fig:order_by_likes} presents the number of likes received by different models. Here we count the likes for each model, where each account can like a model only once, making the statistics a credible measure of individual model popularity. We observe that 7B-scale models are the most popular, offering strong performance while maintaining relatively low resource consumption. Closely following are 70B-scale models, which are highly valued for their exceptional performance.

\textbf{The performance of open-source LLMs have steadily improved, and the size of models for achieving the same performance is shrinking.}
As shown in Figure~\ref{fig:all_average_performance_pre-trained_models_over_time}, the release of ChatGPT spurred a surge of open-source models with rapid performance improvements. These models have increasingly rivaled closed-source counterparts, culminating in Deepseek V3 surpassing GPT-4 on the MMLU benchmark~\citep{liu2024deepseek}. Concurrently, the model size required for comparable performance has steadily decreased; for instance, while a 70B model like LLaMA-2-70B was needed to match GPT-3.5 in July 2023~\citep{touvron2023llama2}, a 9B model such as Yi-1.5-9B was sufficient by May 2024~\citep{young2024yi}.

\textbf{Different Impact of Model Size and Training Strategy on Task Performance.}
To analyze the impact of model size and training strategy, we visualize trends in Figure~\ref{fig:performance_size_train}. We apply equal-frequency binning to handle the skewed size distribution, using the mean score in each bin to represent the central trend and the interquartile range (IQR) to indicate performance variability. The visualization reveals a generally positive correlation between model size and performance, but with a notable performance dip for models in the 10B–20B range. A plausible explanation is that sub-10B models have been extensively optimized, while 10B–20B models lack both the popularity for such optimization and significant scale advantages, thus not reaching their full potential.

Further analysis of specific benchmarks reveals distinct patterns. On the GSM8K benchmark, performance differences across models are more pronounced than on other tasks, highlighting significant disparities in mathematical capability. In contrast, post-training provides the largest gains on TruthfulQA, demonstrating its effectiveness in enhancing factual accuracy.

\begin{figure}[t]
    \centering
    \includegraphics[width=0.48\textwidth]{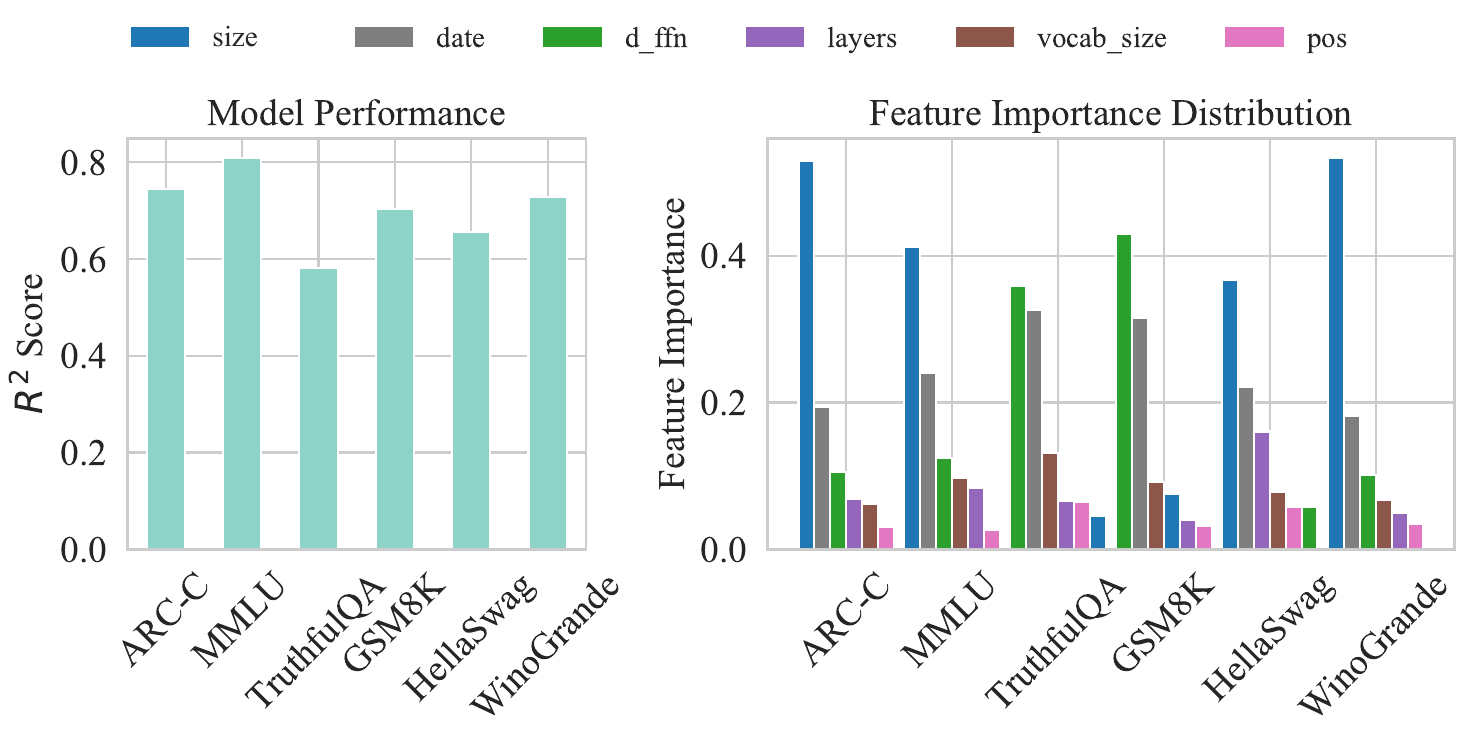}  
    \caption{Regression analysis of key parameters and performance across different benchmarks using the Random Forest algorithm, with corresponding $R^2$ scores and feature importance.}
    \label{fig:model_performance_all_feature_importance}
\end{figure}

\section{Attributing LLMs' Performance to Structure Factors}

\textbf{Scores on ARC-C, HellaSwag, and WinoGrande are highly correlated.} We compute Spearman rank correlation coefficients~\citep{fieller1957tests} to assess performance relationships across datasets (Figure~\ref{fig:performance_spearman_rank_correlation}). This non-parametric metric ranges from –1 to 1, indicating the strength and direction of monotonic associations. The results reveal strong correlations among ARC-C, HellaSwag, and WinoGrande, likely due to their shared focus on reasoning ability.

\textbf{Regression analysis demonstrates a significant correlation between model structure, hyperparameters, and performance.} We aim to explore the relationship between structure, hyperparameters, and the performance of LLMs. To this end, we selected a set of key parameters and employed various machine learning (ML) algorithms for regression analysis to investigate how these parameters correlate with model performance, including Random Forest~\citep{breiman2001random}, Linear Regression, Decision Tree~\citep{quinlan2014c4}, SVR~\citep{cortes1995support}, Ridge~\citep{hoerl1970ridge}, Lasso Regression~\citep{tibshirani1996regression}, $k$-Nearest Neighbors~\citep{kramer2013k}, and Gradient Boosting~\citep{friedman2001greedy}. Especially, we fine-tuned the LLaMA-2-7B model for regression tasks using LLaMA-Factory~\citep{zheng2024llamafactory} and LoRA~\citep{hu2021lora} techniques, employing a text-based format. The detailed experiment configurations of the models used, along with examples of predictions from the fine-tuned LLaMA-2-7B, can be found in Appendix~\ref{app:exp_resource} and Appendix~\ref{app:hyperparameter}. 

We utilize the \( R^2 \) score, also known as the coefficient of determination, to assess the effectiveness of each regression method. \( R^2 \) is given by Equation~\ref{eq1}:
\begin{equation}
  \label{eq1}
    R^2 = 1 - \frac{\sum_{i=1}^{n} (y_i - \hat{y}_i)^2}{\sum_{i=1}^{n} (y_i - \bar{y})^2},
\end{equation}
where \( y_i \) are the actual values, \( \hat{y}_i \) are the predicted values, and \( \bar{y} \) is the mean of the actual values. A higher \( R^2 \) indicates a better fit of the model to the data.

The corresponding \( R^2 \) scores are shown in \autoref{ml_performance}. Machine learning results reveal a clear correlation between model structure and performance, with random forest achieving the highest predictive accuracy. We also compute the Mean Absolute Error (MAE), which remains below 6 for most tasks except GSM8K, indicating practical predictive value. Here, the focus is not on pursuing precise prediction, but on utilizing these results for subsequent analysis—for instance, to assess the relative influence of different structural factors. Given the multifactorial nature of LLM performance, the consistent and significant correlations observed robustly highlight key architectural levers. Moreover, the fine-tuned model can reasonably predict performance across benchmarks using a text-based format, suggesting a future where LLMs autonomously analyze data, adapt structures, and evolve to meet new challenges~\citep{tao2024survey}.

\textbf{Model size and release date are the primary factors influencing performance.} To evaluate the impact of these features, we extracted feature importance from the Random Forest algorithm, which demonstrated the best performance among the tested methods. This feature importance reflects the contribution of each feature in reducing node impurity (measured by mean squared error, MSE) across all tree splits~\citep{genuer2010variable}. 
Formally, the feature importance of feature \( f \) is given by Equation~\ref{eq2}:
\begin{equation}
    \label{eq2}
    I_f = \sum_{t \in T} \Delta \mathrm{Impurity}(t, f),
\end{equation}
where \( T \) represents the set of all decision trees, and \( \Delta \mathrm{Impurity}(t, f) \) denotes the weighted decrease in mean squared error at node \( t \) resulting from the use of feature \( f \) for splitting.

As presented in Figure~\ref{fig:model_performance_all_feature_importance}, we observe that benchmark performance is most strongly correlated with model size and release date. The correlation with model size is relatively straightforward. The release date reflects not only improvements in training techniques but also a steady increase in pre-training token counts: from 1T in LLaMA, to 2T in LLaMA-2, 8T in Mistral~\citep{jiang2023mistral7b}, and roughly 15T in the latest models~\citep{dubey2024llama}.

\begin{table*}[t]
  \centering
  \setlength{\tabcolsep}{5pt}
  \begin{tabular}{l|cccccc}
    \hline
    \textbf{Model} & \textbf{ARC-C} & \textbf{MMLU} & \textbf{TruthfulQA} & \textbf{GSM8K} & \textbf{HellaSwag} & \textbf{WinoGrande} \\
    \hline
    Random Forest         & {\bf 75\%} & {\bf 81\%} & {\bf 58\%} & {\bf 70\%} & {\bf 66\%} & {\bf 73\%} \\
    Linear Regression     & 52\% & 54\% & 32\% & 44\% & 41\% & 50\% \\
    Decision Tree         & 69\% & 79\% & 54\% & 63\% & 57\% & 68\% \\
    SVR                   & 64\% & 68\% & 46\% & 58\% & 51\% & 62\% \\
    Ridge                 & 52\% & 54\% & 32\% & 44\% & 41\% & 50\% \\
    Lasso Regression      & 52\% & 54\% & 32\% & 44\% & 41\% & 50\% \\
    $k$-Nearest Neighbors & 71\% & 77\% & 50\% & 67\% & 62\% & 69\% \\
    Gradient Boosting     & 72\% & 78\% & 56\% & 67\% & 64\% & 71\% \\
    MLP                   & 68\% & 74\% & 49\% & 64\% & 56\% & 66\% \\
    LLM Fine-tune         & 60\% & 65\% & 17\% & 39\% & 51\% & 56\% \\
    \hline
  \end{tabular}
  \caption{\label{ml_performance}
    $R^2$ scores when predicting LLMs' performance across different datasets using key parameters with various methods.
  }
\end{table*}

\begin{figure*}[t]
    \centering
    \begin{subfigure}{0.47\textwidth}
        \centering
        \includegraphics[width=\textwidth]{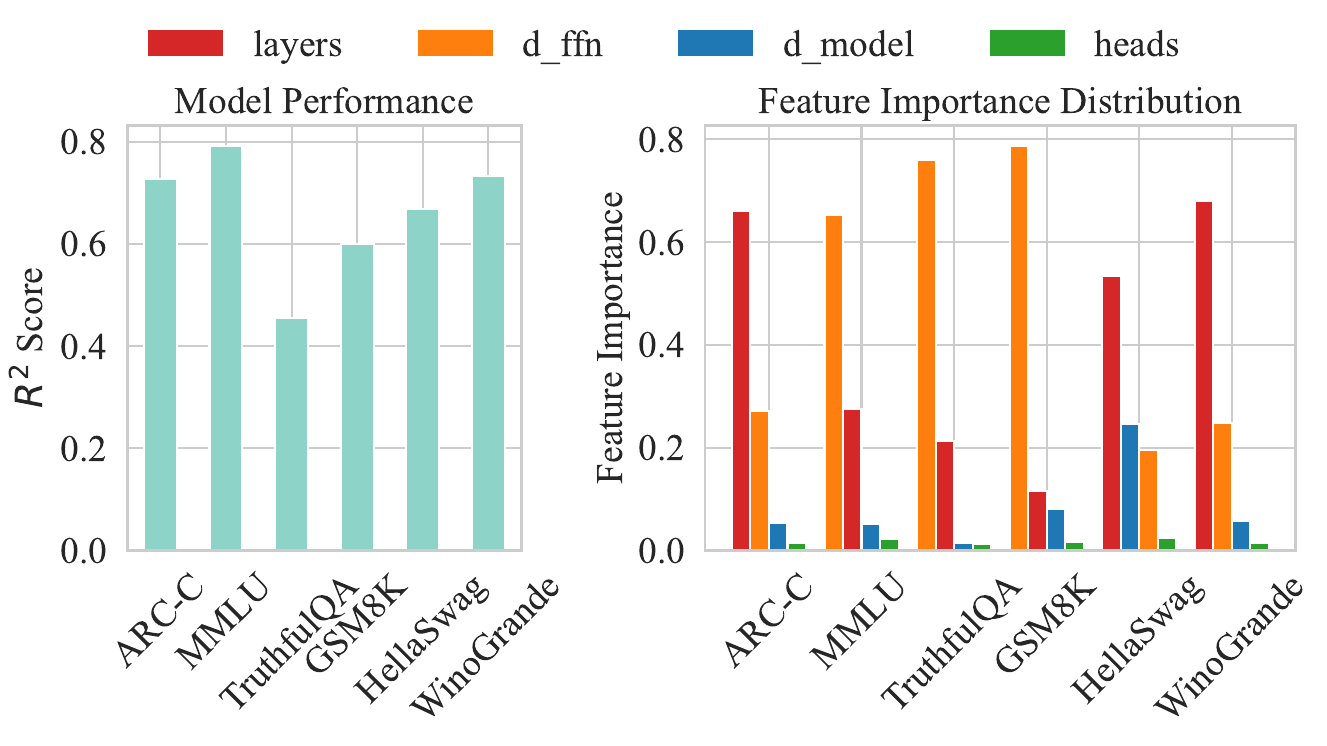}
        \caption{}
        \label{fig:model_performance_feature_importance}
    \end{subfigure}\hfill
    \begin{subfigure}{0.47\textwidth}
        \centering
        \includegraphics[width=\textwidth]{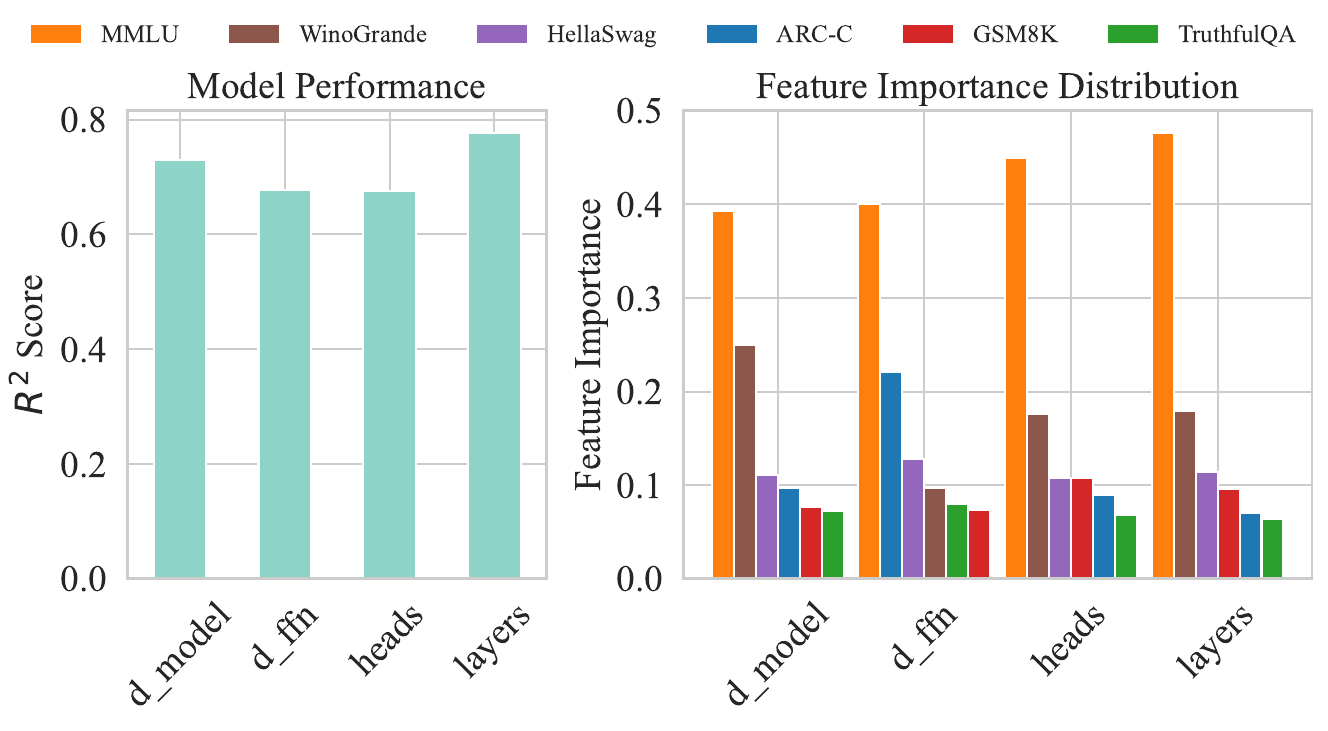}
        \caption{}
        \label{fig:model_architecture_feature_importance}
    \end{subfigure}
    \caption{Regression analysis of model structure and performance using the Random Forest algorithm. (a) Performance prediction from structural parameters, showing layer depth as most influential for reasoning tasks and FFN size for knowledge- and math-oriented tasks. (b) Structure prediction from performance, where MMLU emerges as the most indicative benchmark.}
\end{figure*}

\textbf{Layer depth and \( d_{ffn} \) impact different types of benchmarks.} 
We analyzed key structural variables—\texttt{layers} (layer depth), \texttt{d\_ffn} (FFN intermediate size), \texttt{d\_model} (hidden dimension), and \texttt{heads} (attention heads)—as shown in Figure~\ref{fig:model_performance_feature_importance}. Our results suggest that \texttt{layers} mainly affects reasoning tasks (e.g., ARC-C, HellaSwag, WinoGrande), while \texttt{d\_ffn} more strongly influences mathematical ability and knowledge accuracy, as seen in GSM8K, MMLU, and TruthfulQA. The robustness and generalizability of our findings are further supported by experiments that control for developer proficiency and development timing (Appendix~\ref{app:main_model_regression}).

This aligns with prior analyses: layer depth governs the degree of non-linearity, thereby enhancing reasoning abilities~\citep{jin2024exploring, mueller2023plant, ye2024physics}, whereas empirical studies indicate that LLMs store knowledge mainly in the FFNs~\citep{geva2020transformer, stolfo2023mechanistic}, with larger $d_{ffn}$ substantially boosting memory capacity. This also concurs with findings that increasing the number of experts in MoE models—viewed as an extension of the FFNs—improves performance on knowledge-intensive tasks but not on reasoning~\citep{jelassi2024mixture, fedus2022switch}.

Furthermore, \citet{mirzadeh2024gsm} observe that even minor modifications to the GSM8K dataset cause a significant performance drop, suggesting that such models primarily rely on memorization to solve mathematical problems. Meanwhile, \citet{stolfo2023mechanistic} find that LLMs mainly execute basic arithmetic operations within the FFNs. Together, these studies explain why \( d_{ffn} \) plays a more critical role than layer depth on the GSM8K task.

\textbf{Extending the Analysis to Diverse Tasks and Deployment Metrics.}
To complement our initial analysis on general-purpose benchmarks, we extend the investigation to specialized domains—long-context reasoning, coding, instruction-following, and practical deployment metrics. For this extension, we curate task-specific datasets of relevant models and their performance. Random-forest regression consistently shows that different structural factors dominate distinct capabilities; the corresponding feature-importance scores are summarized in Table~\ref{tab:code_ifeval_feature_improtance}.

On BigCodeBench (coding tasks), the regression achieves $R^2=66.2\%$, with \texttt{layers} emerging as the most influential factor, suggesting that deeper architectures benefit programming-oriented reasoning. In contrast, for IFEval (instruction following; $R^2=48.2\%$), \texttt{d\_ffn} is the dominant contributor. For long-context reasoning on LongBench v2 ($R^2=70.26\%$), \texttt{d\_ffn} overwhelmingly dominates, indicating that wider FFNs are essential for handling extended contexts effectively.

For deployment-related performance using the LLM-Perf Leaderboard, decoding speed regression yields $R^2=81.54\%$, with \texttt{d\_model} and \texttt{d\_ffn} acting as joint primary determinants with near-identical importance. For memory usage ($R^2=88.36\%$), \texttt{d\_model} emerges as the most influential factor.

\begin{table}[h]
    \centering
    \setlength{\tabcolsep}{4pt}
    \begin{tabular}{@{}l|cccc@{}}
        \hline
        Benchmark & layers & d\_model & d\_ffn & heads \\
        \hline
        BigCodeBench & \textbf{35.4\%} & 33.5\% & 23.1\% & 8.1\% \\
        IFEval       & 29.3\% & 25.2\% & \textbf{39.0\%} & 6.5\% \\
        Longbench v2 & 19.3\% & 16.7\% & \textbf{49.9\%} & 14.1\% \\
        Decode Speed & 27.6\% & \textbf{32.6\%} & 32.4\% & 7.5\% \\
        Memory Usage & 10.9\% & \textbf{61.3\%} & 27.4\% & 0.4\% \\
        \hline
    \end{tabular}
    \caption{\label{tab:code_ifeval_feature_improtance} 
     Feature importance of structural variables in random forest regression models across diverse tasks. The results highlight that different tasks exhibit varying sensitivities to different structural parameters.}
\end{table}

\textbf{MMLU is the most representative benchmark.}  
Our analysis reveals that MMLU performance is the key feature for predicting model structure, as shown by the feature importance values in Figure~\ref{fig:model_architecture_feature_importance}. This supports the hypothesis that MMLU scores best capture overall model performance and aligns with how organizations like OpenAI, Anthropic, Mistral, and Qwen typically showcase model capabilities on MMLU.

\section{Mechanistic Interpretability Analysis}
\subsection{Validating the Impact of Layer Depth via Layer Pruning}

We apply the ShortGPT~\citep{men2024shortgpt} method to prune LLaMA-2-7B to validate the impact of layer depth. The experiments on the Qwen-2-7B and LLaMA-2-70B models are shown in Appendix~\ref{app:qwen7b_llama70b_layers_prune}. By pruning a small number of layers with the lowest Block Influence (BI) scores (Equation~\ref{eq3}), we introduce controlled variations in model depth while minimizing disruption to the model’s overall capabilities. This setup enables us to examine how depth adjustments affect performance across different tasks under comparable conditions.
\begin{equation}
    {\rm BI}_i = 1 - {E}_{X,t} \frac{X_{i,t}^T X_{i+1,t}}{\|X_{i,t}\|_2 \|X_{i+1,t}\|_2},
    \label{eq3}
\end{equation}
where $X_{i,t}$ is the $t^{th}$ row of the hidden state at layer $i$. A lower BI score indicates higher cosine similarity between $X_i$ and $X_{i+1}$, suggesting that the layer contributes less transformation and is thus less critical.

By averaging BI scores over multiple benchmarks for the LLaMA-2-7B model, we observe consistent patterns across layers, as shown in Appendix~\ref{app:BI_score}, making it challenging to use BI scores alone to differentiate the functional roles of individual layers across tasks. Therefore, we prune layers 21 through 29, which have the lowest BI scores.

We observe an anomaly in the GSM8K benchmark, which requires models to generate precise numerical answers rather than selecting from multiple choices as in other benchmarks. This unique task structure makes GSM8K not directly comparable to the others. Therefore, we exclude GSM8K from this experiment.

\begin{figure}[t]
    \centering
    \includegraphics[width=0.4\textwidth]{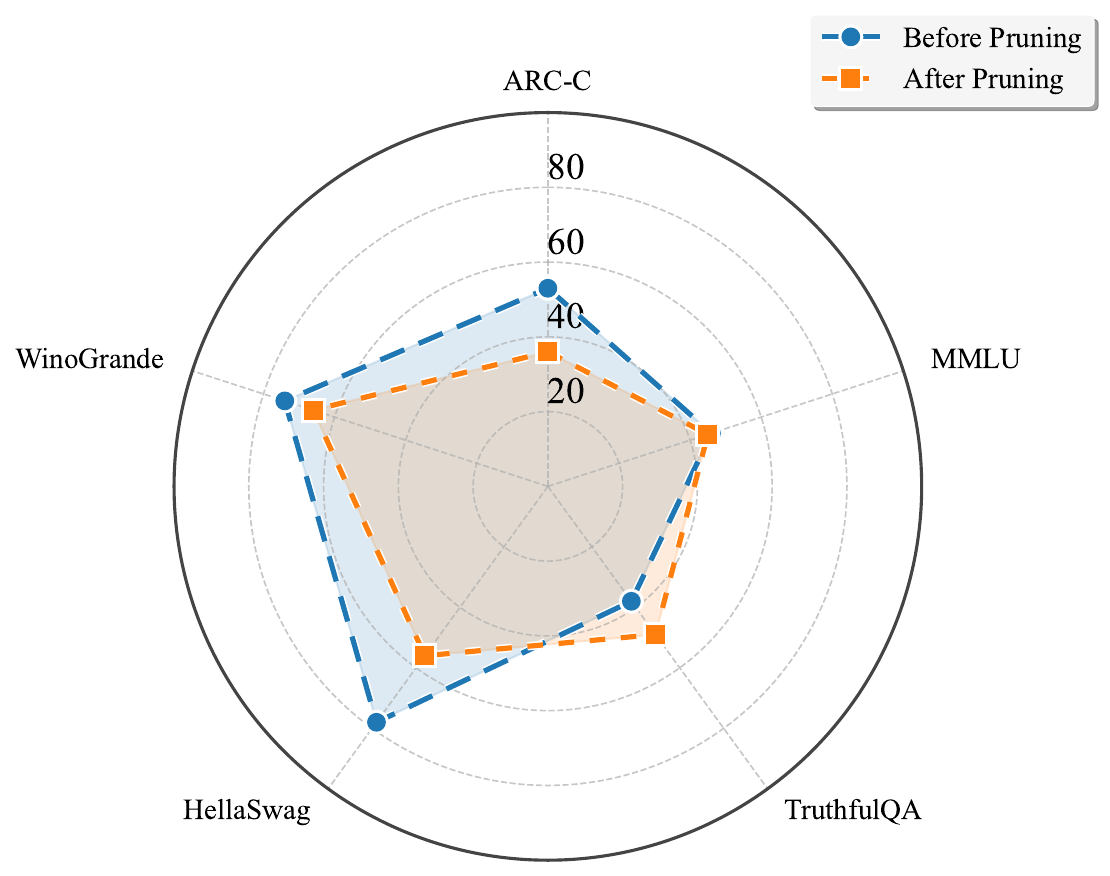}
    \caption{Performance of LLaMA-2-7B before and after pruning layers 21–29. Pruning the least important layers causes a clear drop on reasoning tasks, while the effect on knowledge-focused tasks is minimal, with TruthfulQA even slightly improving—highlighting the critical role of model depth in reasoning ability.}
    \label{fig:llama_enhanced_performance_radar}
\end{figure}

After pruning these layers, we evaluate the model using lm-evaluation-harness~\citep{eval-harness} following \textit{the leaderboard} protocols, comparing its performance before and after pruning across multiple benchmarks. The results are shown in Figure~\ref{fig:llama_enhanced_performance_radar}.

Pruning leads to significant performance drops on benchmarks where layer depth is a critical factor (ARC-C, HellaSwag, WinoGrande), confirming the random forest regression results (Figure~\ref{fig:model_performance_feature_importance}). Conversely, benchmarks less dependent on layer depth (e.g., MMLU, TruthfulQA) show minimal degradation, with TruthfulQA even improving slightly, further validating our analysis.

\begin{figure}[t]
    \includegraphics[width=0.47\linewidth]{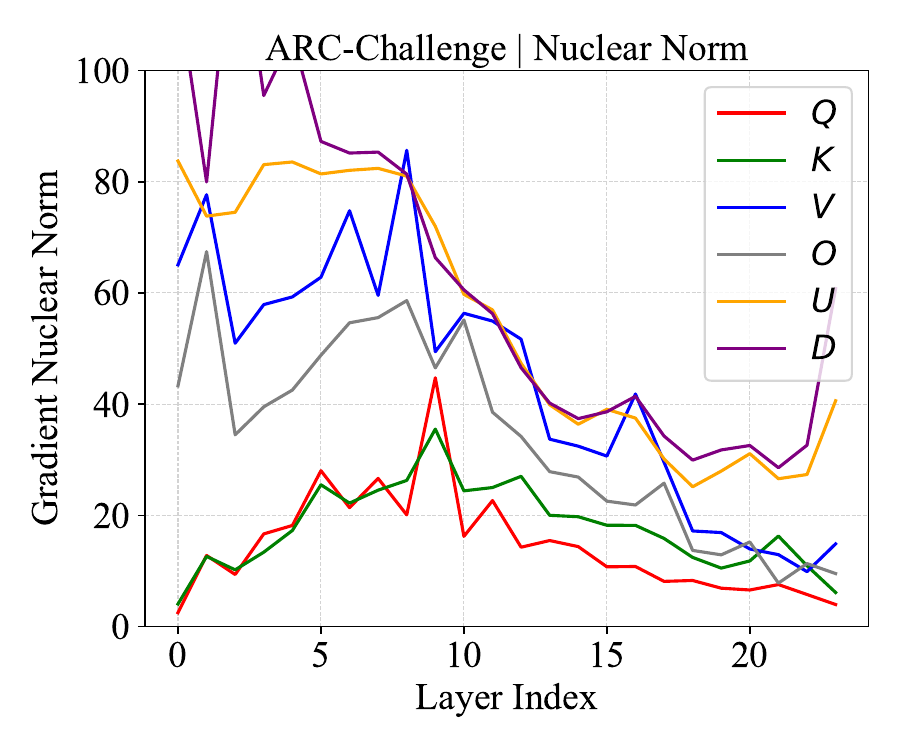} \hfill
    \includegraphics[width=0.47\linewidth]{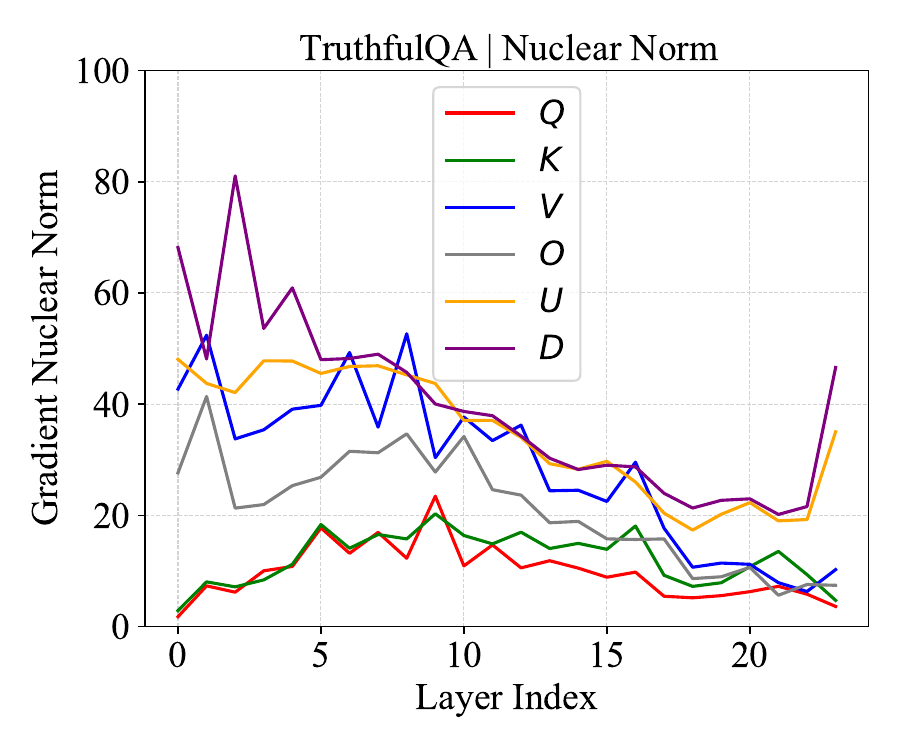}
    \caption{Layer-wise gradient analysis during fine-tuning of Qwen-2-0.5B on the ARC-C and TruthfulQA benchmarks.}
    \label{fig:gradient_truthfulqa_arc}
\end{figure}

\subsection{Validating Findings through Layer-wise Gradient Analysis}
\label{sec:grad_method}
Following the gradient analysis methodology of \citet{DBLP:journals/corr/abs-2410-23743}, we evaluate the gradients during fine-tuning of Qwen-2-0.5B on the ARC-C and TruthfulQA benchmarks, which are representative tasks where layers depth and \( d_{\text{ffn}} \), respectively, are identified as the most influential structural factors.

Our analysis focuses on six major weight matrices in each decoder layer: the Query ($Q$), Key ($K$), Value ($V$), and Output ($O$) projections in the attention module, as well as the Up ($U$) and Down ($D$) projections in the FFN module. We denote $X \in \{Q, K, V, O, U, D\}$.

The loss $L_{\theta}$ corresponds to the cross-entropy loss for next-token prediction used in supervised fine-tuning, where only the response tokens contribute to the overall loss, and instructions are ignored. We perform multiple backward passes until gradients from all entries in the dataset are accumulated.

For the weight matrix $X_i$ of the $i$-th layer and its corresponding gradient $G_{X,i}$, we measure the concentration of its gradient spectrum on dominant singular values using the Nuclear Norm $s_{X,i}$. This provides insights into the gradient behavior across different layers and tasks. The Nuclear Norm is given by Equation~\ref{eq4}:

\begin{equation}
    s_{X, i} = \| G_{X, i} \|_* = \sum_{j=1}^{{\rm min} (m, n)}{|\sigma_j|},
    \label{eq4}
\end{equation}
where $\sigma_j$ denotes the $j$-th singular value, computed via singular value decomposition (SVD), as shown in Equation~\ref{eq5}:

\begin{equation}
\begin{aligned}
    & \Sigma = {\rm diag}\left( \sigma_1, \sigma_2, \cdots, \sigma_{{\rm min}(m, n)} \right), \\
    & G_{X, i} = U \Sigma V^{\top}.
\end{aligned}
\label{eq5}
\end{equation}

The results of this analysis are shown in Figure~\ref{fig:gradient_truthfulqa_arc}. We observe that gradients in the deeper layers of the ARC-C benchmark remain relatively high, indicating that deeper layers play a more critical role in successfully completing reasoning tasks. This finding aligns with our earlier observation that layer depth is the key structural factor for ARC-C. In contrast, gradients in the deeper layers of the TruthfulQA benchmark are substantially lower, suggesting that these layers contribute less to this memory-centric task. 

The experiment on LLaMA-3.2-3B is presented in Appendix~\ref{app:llama3b_layer_grad}. Meanwhile, a deeper investigation into the gradient dynamics, as detailed in Appendix~\ref{app:layer_grad}, further supports this hypothesis.

\section{Related Work} 
\subsection{Model Evaluation} 
In the field of LLMs, evaluating and comparing model performance is crucial for advancing technology. One of the most prominent platforms for benchmarking is the Open LLM Leaderboard~(\textit{the leaderboard}, \citealp{open-llm-leaderboard, open-llm-leaderboard-v2}), hosted by HuggingFace, which provides a standardized environment for evaluating various large-scale models across numerous tasks. 

Although \textit{the leaderboard} provides practical performance comparisons between LLMs, it overlooks the structural configurations of the models. There has been limited exploration of the relationships between these configurations and the performance across different datasets. Our work aims to address this gap by combining model structural configurations with performance data from \textit{the leaderboard}. This additional dimension provides valuable insights into how model structure affects performance, complementing the benchmark scores.

\subsection{Mechanistic Interpretability} 
Mechanistic interpretability (MI)~\citep{olah2020zoom, sharkey2025open} is an emerging subfield of interpretability that aims to understand a neural network model by reverse-engineering its internal computations. Recently, MI has garnered significant attention for interpreting transformer-based LLMs, showing promise in providing insights into the functions of various model components (e.g., neurons, attention heads), offering mechanistic explanations for different model behaviors, and enabling users to optimize the utilization of LLMs \citep{rai2024practical, luo2024understanding, zhao2024towards, yao2025deep}.

However, most research on MI has focused on specific components or specialized tasks, without providing a unified explanation of how the overall structure of LLMs relates to their general capabilities. In contrast, our study adopts a data-driven approach: first, by uncovering phenomena through mining structured datasets, and then applying MI techniques to validate these phenomena, we aim to achieve a comprehensive understanding of how model structures and performance interact.

\section{Conclusion}
This study provides a comprehensive, data-driven analysis of LLMs through a large-scale dataset that captures structural configurations and their performance across diverse benchmarks. By systematically tracing the evolution of LLMs, we identify emerging trends and offer insights into future directions. Our findings underscore the critical influence of structural configurations on model performance, validated through mechanistic interpretability techniques. This work delivers actionable, data-driven guidance for optimizing LLM design, paving the way for the development of more efficient, scalable, and adaptable models to meet the demands of diverse real-world applications.

\section*{Acknowledgements}
This work was supported by the National Natural Science Foundation of China under Grant No. 62306216, Natural Science Foundation of Hubei Province of China Grant No.2023AFB816. Hai Zhao was funded by The Major Program of Chinese National Foundation of Social Sciences under Grant ‘The Challenge and Governance of Smart Media on News Authenticity’ [No. 23\&ZD213].

\section*{Limitations}
This study focused on a specific set of tasks, potentially limiting the generalizability of our findings. Different applications may involve distinct requirements and data characteristics. Future work should explore a broader range of tasks to improve the robustness and applicability of our conclusions.

Our mechanistic interpretability analysis was limited to methods such as layer pruning and gradient analysis. While these techniques provided valuable insights, they may not fully capture the complex internal dynamics of LLMs. Future research could incorporate a wider variety of interpretability tools to validate and complement our findings, thereby offering a more comprehensive understanding of model behavior.

\section*{Ethics Statement}
All training and evaluation datasets used in this study are publicly available under open-access licenses and intended solely for research purposes. These datasets contain no personal or identifiable information, nor any offensive content. The data analyzed in this work pertains exclusively to model structure and performance metrics.

All datasets developed or used in this research will be released under the MIT License. We share these resources to promote transparency, reproducibility, and further research within the community. We encourage others to build upon and improve our work, provided they adhere to the terms of the MIT License.

\bibliography{acl}
\clearpage
\onecolumn
\appendix
\begin{center}
    {\Large\textbf{Appendices}}
\end{center}
\section{Details of the LLMs Structure and Performance Dataset}
\label{app:Details}
\subsection{Detailed Description of Each Column}
\label{app:desc}
As shown in Table~\ref{tab:desc}, each column presents key metrics and attributes of the model, offering valuable insights into characteristics such as its size, structure, and usage statistics.

\begin{table}[h]
    \centering
    \begin{tabular}{c|p{2cm}cp{10cm}}
    \hline
    Column & Name & Unit & Description \\
    \hline
    size        & Model Size           & Billions & 
    The overall parameter count of the model. \\ \hline
    d\_model    & Hidden Dim           & $1$      & 
    The size of the hidden state of the model. Usually describing how wide the model is. \\ \hline
    d\_ffn      & Intermediate Size    & $1$      &
    The size of the intermediate state of the MLP (or GLU) in the FFN of each Transformer Decoder Layer. A wider model usually has a larger d\_ffn.\\ \hline
    heads       & Attention Head Count & $1$      &
    The number of attention heads. \\ \hline
    layers      & Decoder Layer Count  & $1$      &
    The number of Decoder layers. A deeper model is whose layer count is larger.\\ \hline
    kv\_heads   & KV Head Count        & $1$      &
    The number of KV heads. Related with GQA (MQA) and the size of KV cache per token. Equal to the heads count for MHA, 4 to 16 times smaller for GQA variant.\\ \hline
    vocab\_size & Vocabulary Size      & $1$      & 
    The available token count of the tokenizer, as well as the embedding and LM\_head component of the base model. Larger vocab means less sequence length, more efficient in inference but at the cost of more parameter.\\ \hline
    pos         & Maximum Input Position & $1$      &
    The maximum capable input sequence length. Relate with sin and cos value caching of Rotary Positional Embedding, also indicating the long context ability with the model. \\ \hline
    downloads   & Download Count       & $1$      &
    The download count on Hugging Face model pages, reflecting actual usage and interest from the community. \\ \hline
    likes       & Like Count           & $1$      & 
    Users' like count on Hugging Face model pages, reflecting community recognition. \\ 
    \hline
    \end{tabular}
    \caption{Description of each column from our LLMs Structure and Performance Dataset.}
    \label{tab:desc}
\end{table}
\subsection{The example of the LLMs Structure and Performance Dataset}
As shown in Table~\ref{tab:example}, the structure parameters of several models and their performance across different benchmarks are presented, including LLaMA-3-8B, Bloom~\citep{le2023bloom}, Mixtral-8x7B, LLaMA-2-7B, and Mistral-7B.
\begin{table}[ht]
  \centering
  \setlength{\tabcolsep}{10pt} 
  \begin{tabular}{l|ccccc}
    \hline
    \textbf{Parameter} & \textbf{LLaMA-3-8B} & \textbf{bloom} & \textbf{Mixtral-8x7B} & \textbf{LLaMA-2-7B} & \textbf{Mistral-7B}\\
    \hline
    \textbf{size} & 8 & 176 & 46 & 7 & 7\\
    \textbf{d\_model} & 4096 & 14336 & 4096 & 4096 & 4096\\
    \textbf{d\_ffn} & 14336 &  & 14336 & 11008 & 14336\\
    \textbf{heads} & 32 & 112 & 32 & 32 & 32\\
    \textbf{layers} & 32 & 70 & 32 & 32 & 32\\
    \textbf{kv\_heads} & 8 &  & 8 & 32 & 8\\
    \textbf{vocab\_size} & 128256 & 250880 & 32000 & 32000 & 32000\\
    \textbf{pos} & 8192 &  & 32768 & 4096 & 32768\\
    \textbf{likes} & 4883 & 4632 & 3920 & 3633 & 3259\\
    \textbf{downloads} & 556210 & 28821 & 2911366 & 927400 & 3147345\\
    \textbf{ARC-C} & 60.24 & 50.43 & 66.38 & 53.07 & 59.98\\
    \textbf{HellaSwag} & 82.23 & 76.41 & 86.46 & 78.59 & 83.31\\
    \textbf{MMLU} & 66.7 & 30.85 & 71.88 & 46.87 & 64.16\\
    \textbf{TruthfulQA} & 42.93 & 39.76 & 46.81 & 38.76 & 42.15\\
    \textbf{WinoGrande} & 78.45 & 72.06 & 81.69 & 74.03 & 78.37\\
    \textbf{GSM8K} & 45.19 & 6.9 & 57.62 & 14.48 & 37.83\\
    \hline
  \end{tabular}
  \caption{\label{tab:example}
    Examples from our LLMs Structure and Performance Dataset.
  }
\end{table}

\section{Experimental Details} \label{app:exp}
\subsection{Resources Used in the Experiments} \label{app:exp_resource}
All experiments utilized a total of 200 GPU hours. The tasks included regression analysis of model structure and performance, fine-tuning the LLaMA-2-7B model for regression tasks using the Low-Rank Adaptation (LoRA) technique and the LLaMA-Factory framework\footnote{\url{https://github.com/hiyouga/LLaMA-Factory}}, pruning specific layers of the LLaMA-2-7B model, and evaluating the model on ARC-C, TruthfulQA, WinoGrande, HellaSwag, and MMLU benchmarks using the lm-evaluation-harness\footnote{\url{https://github.com/EleutherAI/lm-evaluation-harness}}. Additionally, we performed gradient analysis during the fine-tuning of the Qwen-2-0.5B model on the ARC-C and TruthfulQA benchmarks.

\subsection{Hyperparameter Configuration for Regression Models} \label{app:hyperparameter}
For regression analysis of model structure and performance, various models were employed. The hyperparameter configurations for these models are provided in \autoref{model_params}.

The LLaMA-2-7B model was fine-tuned using a text-based format, where the model takes a different structure as input and predicts performance across multiple datasets. As shown in \autoref{llama2_7b_performance_prediction}, the fine-tuned model demonstrates strong performance in accurately predicting outcomes in the specified text format.

\setlength{\tabcolsep}{10pt} 
\begin{table}[ht]
  \centering
  \setlength{\tabcolsep}{10pt} 
  \begin{tabular}{l|l}
    \hline
    \textbf{Model} & \textbf{Hyperparameters} \\
    \hline
    Random Forest              & random\_state=42, n\_estimators=100, max\_depth=None \\
    Linear Regression          & fit\_intercept=True, normalize=False \\
    Decision Tree              & random\_state=42, max\_depth=None, min\_samples\_split=2 \\
    SVR                        & kernel=rbf, C=1.0, epsilon=0.1 \\
    Ridge                      & alpha=1.0, fit\_intercept=True \\
    Lasso Regression           & alpha=0.1, max\_iter=1000 \\
    $k$-Nearest Neighbors      & n\_neighbors=5, algorithm=auto \\
    Gradient Boosting          & n\_estimators=100, learning\_rate=0.1, max\_depth=3 \\
    XGBoost                    & objective=reg:squarederror, n\_estimators=100, learning\_rate=0.1 \\
    MLP                        & hidden\_layer\_sizes=(32, 64, 32), max\_iter=100, activation=relu \\
    LLM Fine-tune              & lora\_target=all, learning\_rate=1.0e-4, num\_train\_steps=3500 \\
    \hline
  \end{tabular}
  \caption{\label{model_params}
    Regression models and their key hyperparameters.
  }
\end{table}

\begin{figure}[ht]
    \centering
    \begin{tcolorbox}[colback=blue!5!white,colframe=blue!75!black,title=Examples of Performance Regression Prediction using Fine - tuned LLaMA-2-7B Model,width=0.95\textwidth]
    \textbf{Prompt1:} You are an AI model expert. Analyze the model structure and predict performance metrics. Model Architecture: Num attention heads: 32, Num hidden layers: 32, Vocab size: 32000, Max position embeddings: 32768, Year: 2024, Month: 1, Day: 3, Model dimension: 4096, FFN hidden dimension: 14336, Model parameters: 7.000B \\
    \textbf{Truth1:} \\ Prediction: ARC-C: \textcolor{green}{55.20}, HellaSwag: \textcolor{green}{78.22}, MMLU: \textcolor{green}{50.30}, TruthfulQA: \textcolor{green}{57.08}, WinoGrande: \textcolor{green}{73.24}, GSM8K: \textcolor{green}{11.45} \\
    \textbf{Answer1:} \\ Prediction: ARC-C: \textcolor{orange}{67.41}, HellaSwag: \textcolor{orange}{86.78}, MMLU: \textcolor{orange}{64.07}, TruthfulQA: \textcolor{orange}{67.68}, WinoGrande: \textcolor{orange}{81.61}, GSM8K: \textcolor{orange}{59.74} \\
    \\
    \textbf{Prompt2:} You are an AI model expert. Analyze the model architecture and predict performance metrics. Model Architecture: Num attention heads: 40, Num hidden layers: 36, Vocab size: 50688, Max position embeddings: 2048, Year: 2023, Month: 2, Day: 27, Model dimension: 5120, FFN hidden dimension: 20480, Model parameters: 12.000B \\
    \textbf{Truth2:} \\  Prediction: ARC-C: \textcolor{green}{41.38}, HellaSwag: \textcolor{green}{70.26}, MMLU: \textcolor{green}{25.63}, TruthfulQA: \textcolor{green}{33.00}, WinoGrande: \textcolor{green}{66.46}, GSM8K: \textcolor{green}{1.44} \\
    \textbf{Answer2:} \\ Prediction: ARC-C: \textcolor{orange}{46.42}, HellaSwag: \textcolor{orange}{70.00}, MMLU: \textcolor{orange}{26.19}, TruthfulQA: \textcolor{orange}{39.19}, WinoGrande: \textcolor{orange}{62.19}, GSM8K: \textcolor{orange}{0.61} \\
    \end{tcolorbox}
    \caption{Performance prediction examples using a fine-tuned LLaMA-2-7B model.}
    \label{llama2_7b_performance_prediction}
\end{figure}

\newpage
\section{Further Experiment Result}
\label{app:further_experiment}
\subsection{Analyzing the Impact of Developer Proficiency and Development Timing}
\label{app:main_model_regression}
The central goal of our study is to uncover unified relationships between model structure and performance through large-scale data mining over structural datasets. Due to the breadth and diversity of our dataset, we expect that secondary factors exert minimal influence on the extracted conclusions, as core patterns can be robustly identified across a wide range of models.

Nevertheless, to ensure that our experimental conclusions are not affected by differences in the development proficiency of various model providers, and to mitigate the possibility that our analysis is overly skewed toward LLaMA-based models, we aimed to achieve broader model representation beyond LLaMA-based architectures while maintaining high model quality.

To this end, we selected models from Hugging Face’s \texttt{open-llm-leaderboard/official-providers} (e.g., LLaMA, MistralAI, DeepSeek, Qwen), which are known to follow high-quality training standards. This filtering process resulted in a dataset where LLaMA-based models and their variants comprised only 27\% of the total, effectively reducing potential bias due to their overrepresentation.

\begin{figure*}[ht]
    \centering
    \begin{subfigure}{0.47\textwidth}
        \centering
        \includegraphics[width=\textwidth]{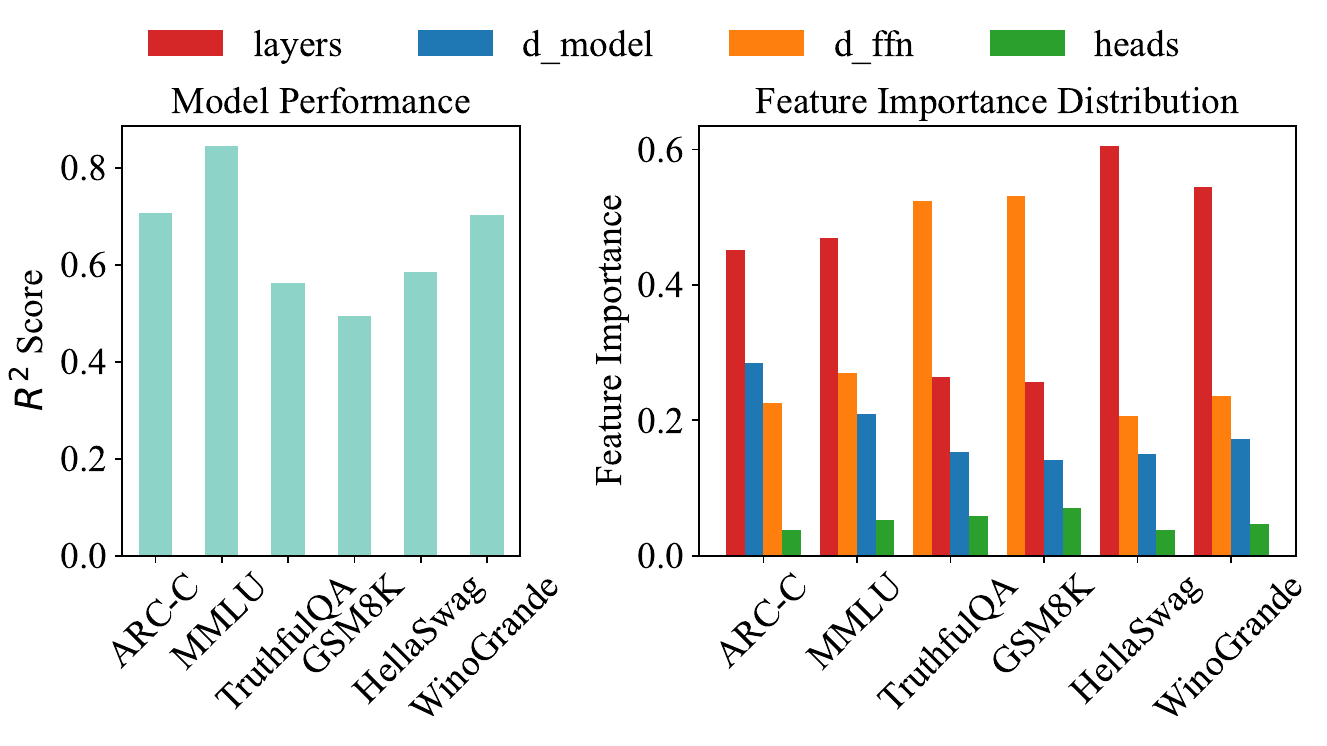}
        \caption{}        \label{fig:main_model_performance_feature_importance}
    \end{subfigure}\hfill
    \begin{subfigure}{0.47\textwidth}
        \centering
        \includegraphics[width=\textwidth]{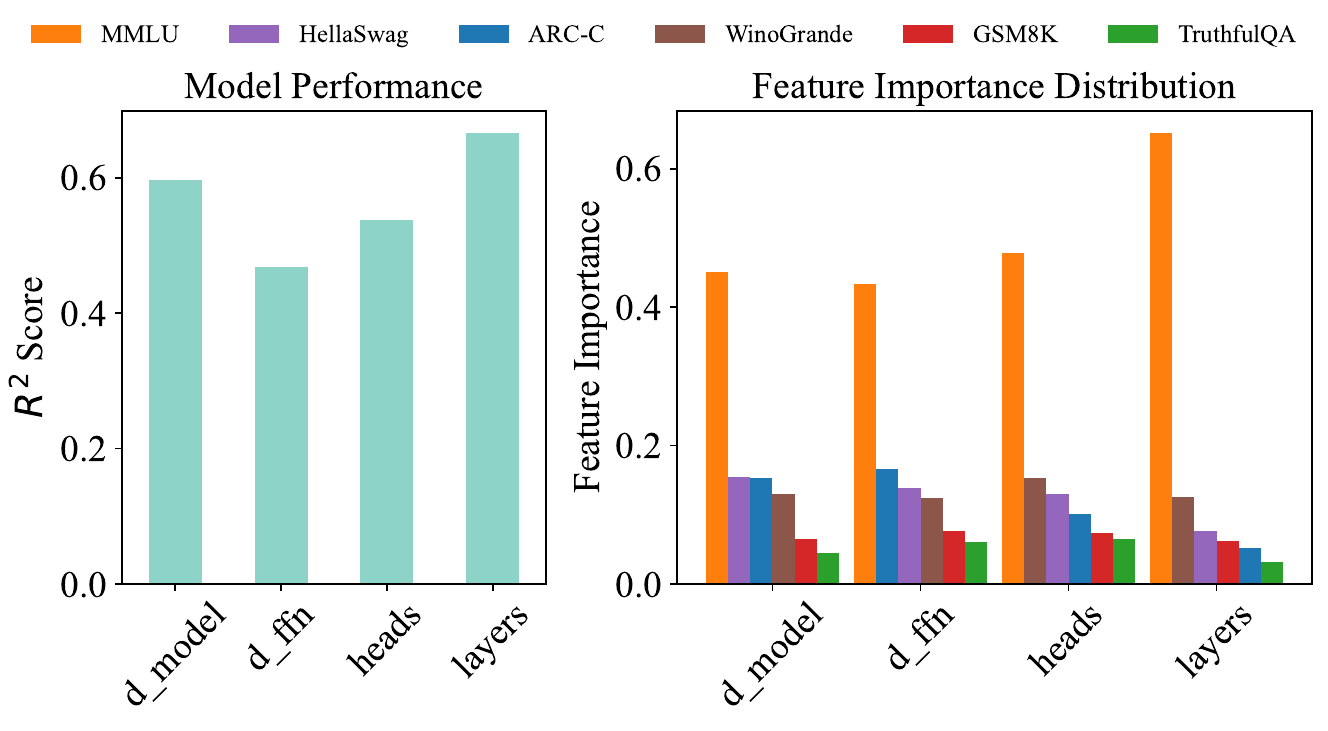}
        \caption{}        \label{fig:main_model_architecture_feature_importance}
    \end{subfigure}
    \caption{Regression analysis of major high-quality model structure parameters and their performance across benchmarks using the Random Forest algorithm. 
    (a) Predicting performance from model structure; 
    (b) Predicting model structure from performance.}
\end{figure*}

As shown in Figure~\ref{fig:main_model_performance_feature_importance}, our results remained consistent with earlier findings: layer depth emerged as the most important structural parameter for ARC-C, HellaSwag, and WinoGrande, while $d_\mathrm{ffn}$ was most critical for TruthfulQA and GSM8K. MMLU was the only exception, likely due to data sparsity.

Meanwhile, as shown in Figure~\ref{fig:main_model_architecture_feature_importance}, performance on the MMLU dataset was identified as the most important parameter for predicting the model’s architectural configuration, which aligns with previous conclusions.

\begin{figure}[ht]
    \centering
    \includegraphics[width=0.6\textwidth]{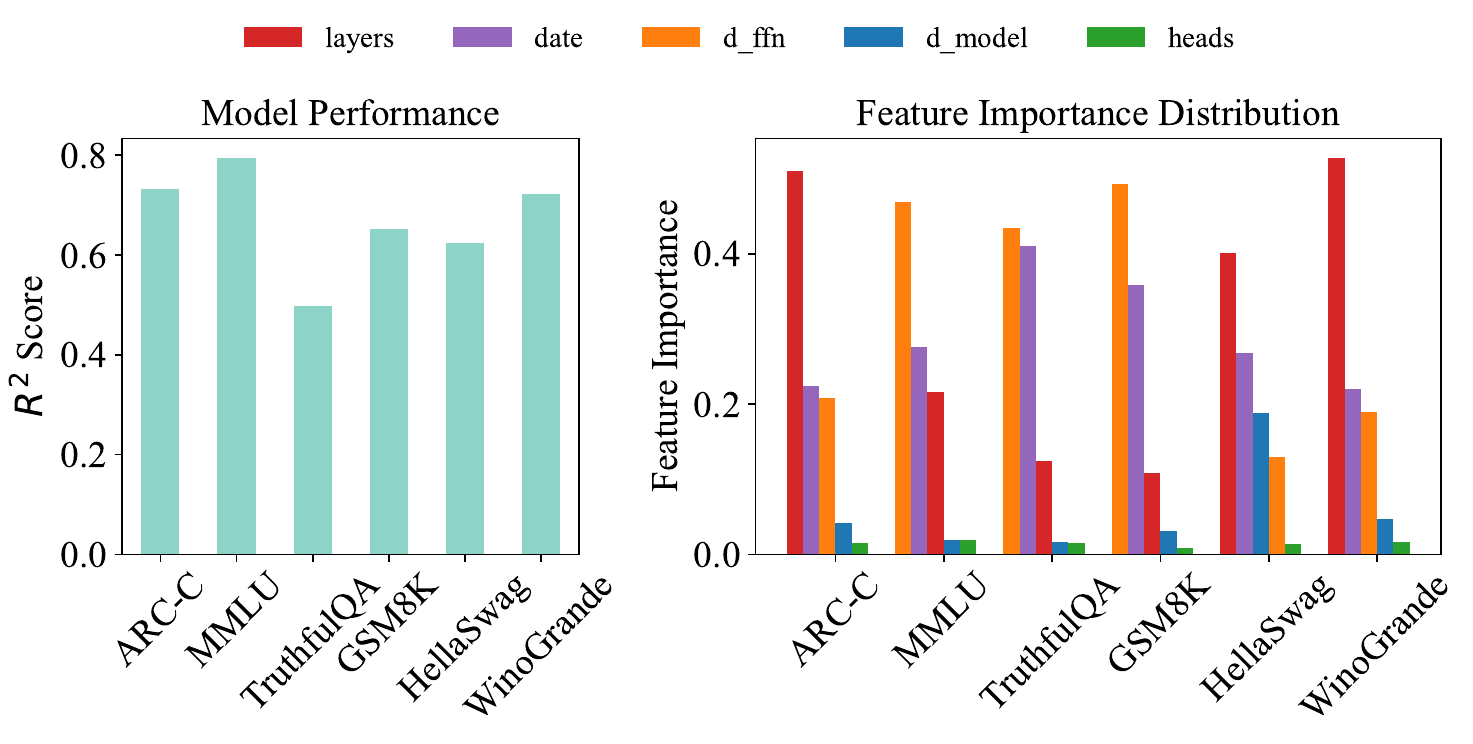}
    \caption{Feature importance in the Random Forest model with date included. Structural features like depth and d\_ffn remain dominant despite temporal effects.}    \label{fig:model_performance_feature_importance_with_date}
\end{figure}

To avoid the impact of temporal variations, we augmented our Random Forest regression model with the date variable. As shown in Figure~\ref{fig:model_performance_feature_importance_with_date}, the resulting $R^2$ scores and feature importance indicate that structural features continue to be significant even when accounting for temporal effects, supporting our conclusion that benchmarks like ARC-C, HellaSwag, and Winogrande rely heavily on model depth. In contrast, $d_\mathrm{ffn}$ emerges as the dominant factor for MMLU, GSM8K, and TruthfulQA.

\subsection{Analysis of BI Scores Across Layers in the LLaMA-2 7B Model across Different Benchmarks}
\label{app:BI_score}
\begin{figure}[ht]
    \centering
    \includegraphics[width=0.8\textwidth]{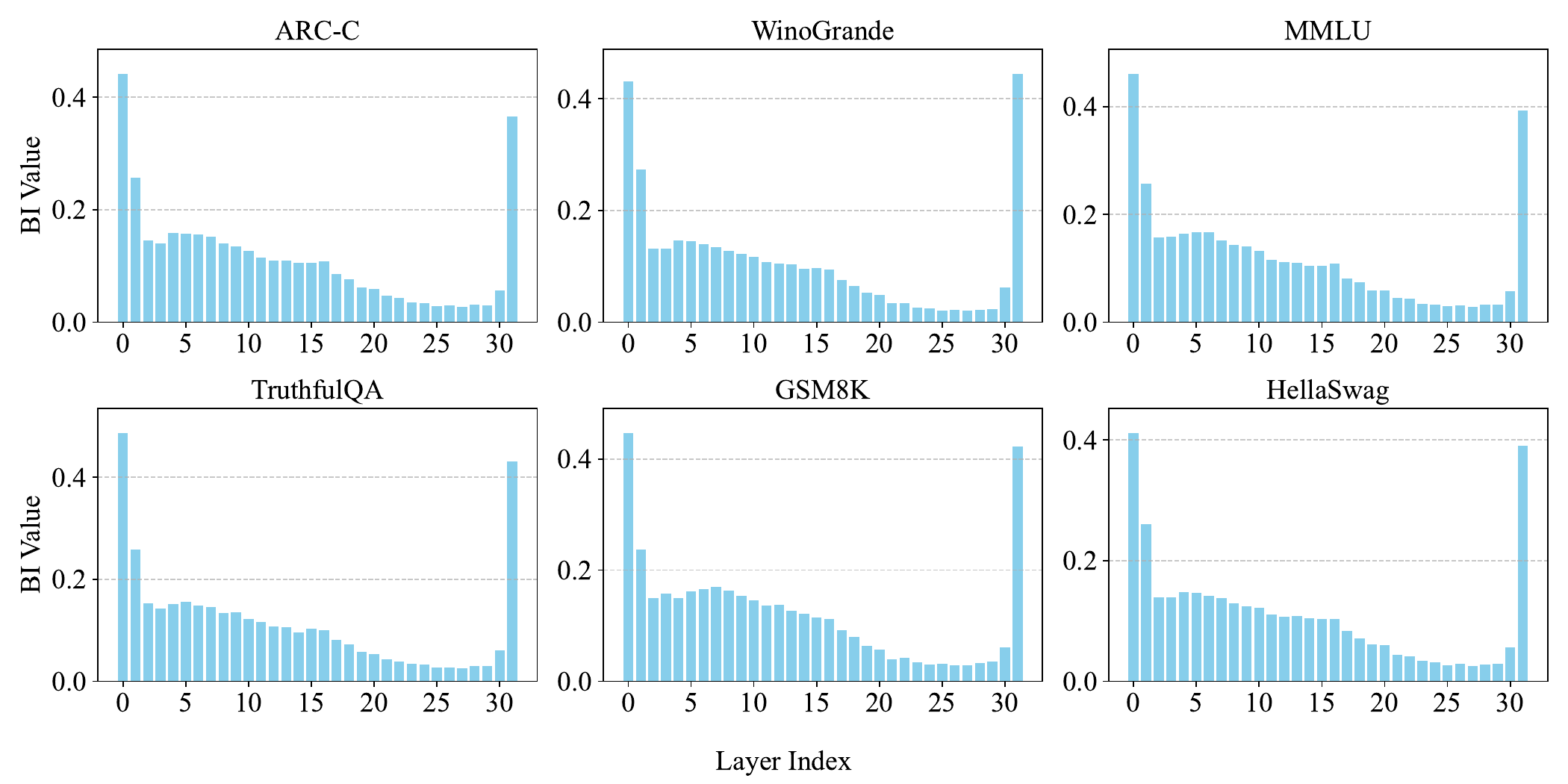}
    \caption{BI scores of different layers in the LLaMA-2-7B model across various benchmarks.}
    \label{fig:combined_bi_visualization}
\end{figure}

As shown in Figure~\ref{fig:combined_bi_visualization}, we present the BI scores for different layers of the LLaMA-2-7B model across various benchmarks. The analysis highlights the relative contribution of each layer to model performance on tasks from diverse domains.

\subsection{Layer Pruning Analysis with Qwen-2-7B and LLaMA-2-70B} 
\label{app:qwen7b_llama70b_layers_prune}

\begin{figure}[t]
    \centering
    \begin{subfigure}[t]{0.48\textwidth}
        \centering
        \includegraphics[width=\textwidth]{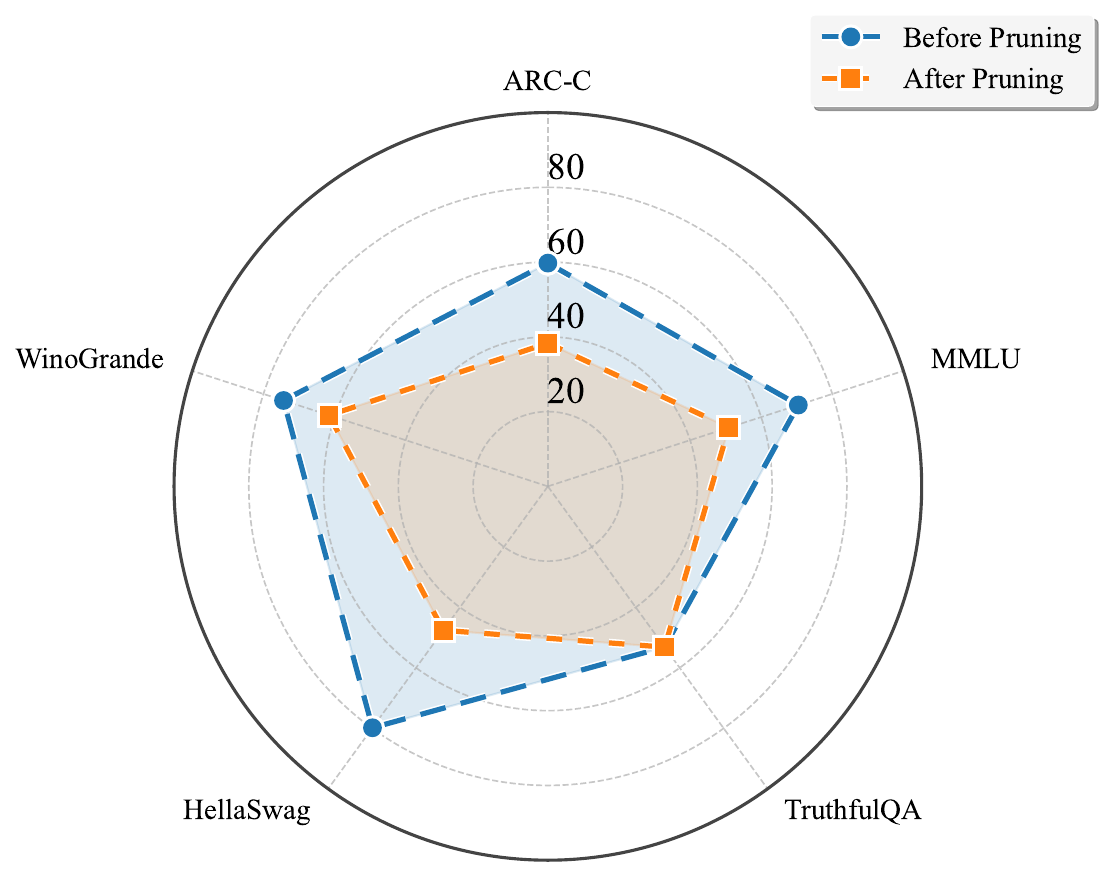}
        \caption{Performance of Qwen-2-7B before and after pruning layers 21–25.}
    \end{subfigure}
    \hfill
    \begin{subfigure}[t]{0.48\textwidth}
        \centering
        \includegraphics[width=\textwidth]{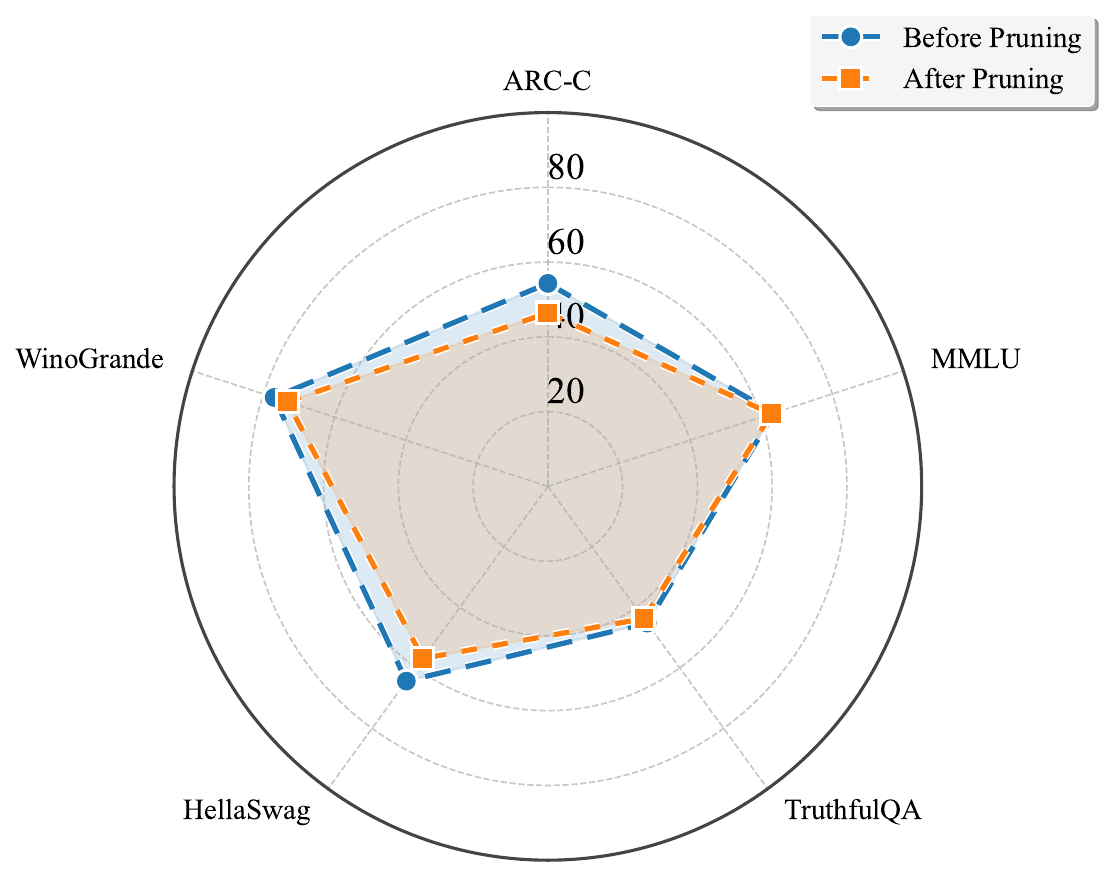}
        \caption{Performance of LLaMA-2-70B before and after pruning layers 58–73.}
    \end{subfigure}
    \caption{Performance across benchmarks before and after pruning. Depth-sensitive tasks (e.g., ARC-C, HellaSwag, WinoGrande) show larger degradation, while others (e.g., MMLU, TruthfulQA) remain relatively stable.}
    \label{fig:pruning_radar}
\end{figure}

To further validate and test the generalizability of our findings from the LLaMA-2-7B experiments, we extended our layer pruning analysis to different model architectures and scales, specifically Qwen-2-7B and a quantized version of LLaMA-2-70B. The results were highly consistent across all models. For Qwen-2-7B, as shown in Figure~\ref{fig:pruning_radar}, pruning led to substantial degradation on depth-sensitive benchmarks (e.g., ARC-C, HellaSwag, WinoGrande), while tasks less dependent on depth (e.g., MMLU, TruthfulQA) exhibited only minor drops. Similarly, for the 80-layer LLaMA-2-70B, we applied the ShortGPT method (Section~6.1) to remove layers 58–73 with low Block Influence (BI) scores. The evaluation results mirrored those of the smaller models: depth-sensitive tasks suffered clear declines, whereas others remained relatively stable. These findings reinforce our conclusion that downstream tasks vary in their sensitivity to model depth.

\subsection{Layer-wise Gradient Analysis with LLaMA-3.2-3B} 
\label{app:llama3b_layer_grad}
\begin{figure}[t]
    \centering
    \includegraphics[width=0.37\linewidth]{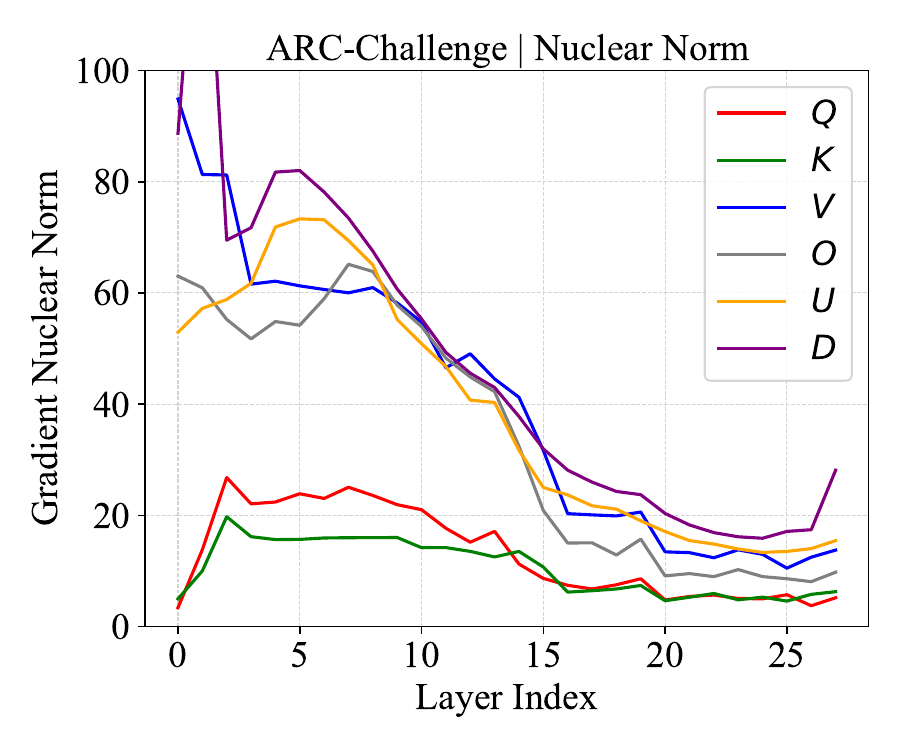}
    \includegraphics[width=0.37\linewidth]{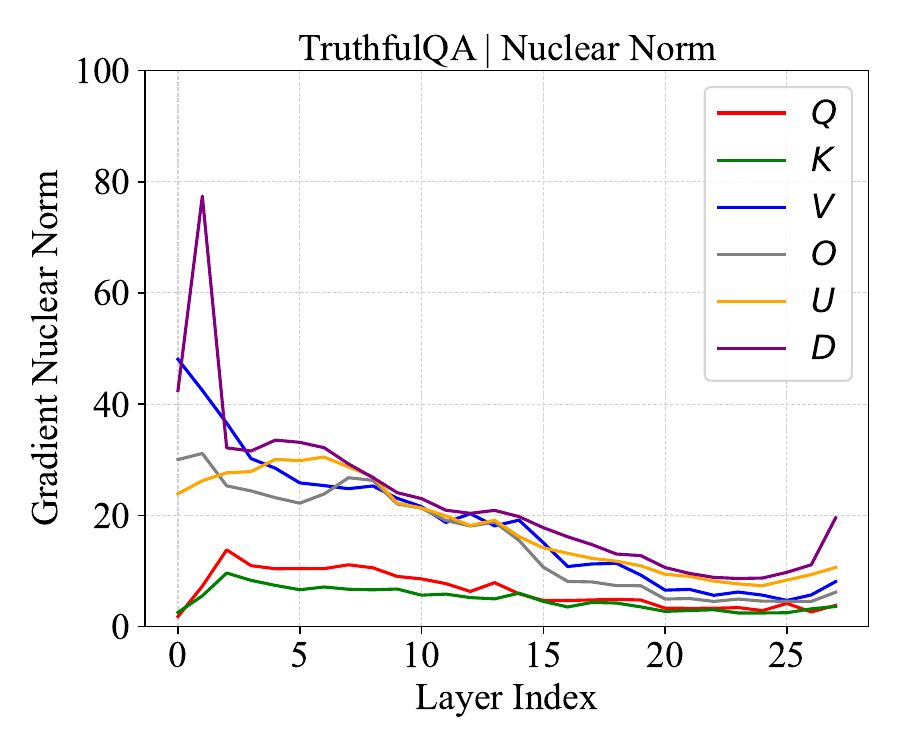}
    \caption{Layer-wise gradient analysis during fine-tuning of  LLaMA-3.2-3B on the ARC-C and TruthfulQA benchmarks.}
    \label{fig:llama_gradient_truthfulqa_arc}
\end{figure}

Similar to the layer-wise gradient analysis conducted on Qwen-2-0.5B, we performed the same experiment on LLaMA-3.2-3B, as shown in Figure~\ref{fig:llama_gradient_truthfulqa_arc}, and found results consistent with our original conclusions. We observe that gradients in the deeper layers of the ARC-C benchmark remain relatively high, while gradients in the deeper layers of the TruthfulQA benchmark are substantially lower. These results further support our previous conclusions.

\subsection{Layer-wise Gradient Analysis with Different Language Styles} \label{app:layer_grad}

\begin{figure}
    \centering
    \includegraphics[width=0.9\linewidth]{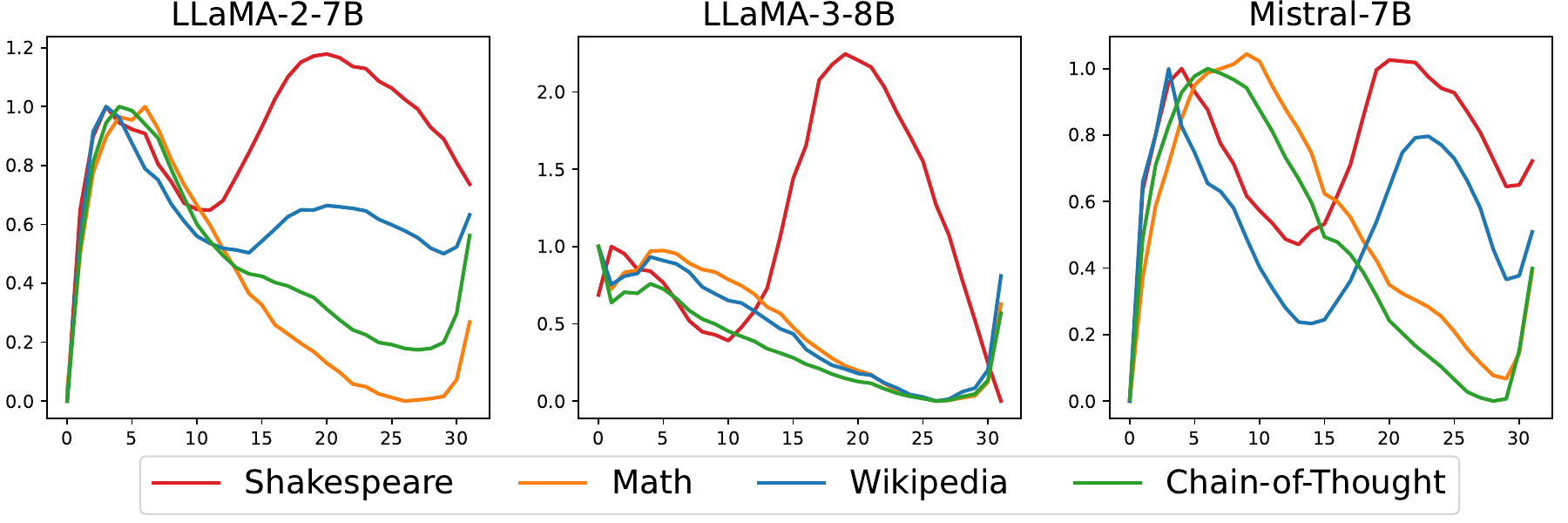}
    \caption{Layer-wise gradient on different corpuses.}
    \label{fig:layer_gradients_corpa}
\end{figure}

We further explore the dynamics of different layers within the model, particularly the deeper layers, to explain how task dependencies vary with model depth. Following the methodology in Section~\ref{sec:grad_method}, we conducted gradient analysis across different corpora. Our findings, shown in Figure~\ref{fig:layer_gradients_corpa}, reveal a significant increase in gradients within the deeper FFN layers when the model encounters distinct linguistic styles or archaic texts. In contrast, for corpora such as plain text or mathematical data, these layers do not exhibit such anomalous gradient behavior.

We observed that the layers responsible for generating the additional gradient peaks largely correspond to the layers excluded in the previous section. Larger gradients typically suggest insufficient training of the corresponding model components. This implies that layers with large gradients in LLMs process language-form-related components, rather than knowledge components abstracted from linguistic forms. In other words, the increased gradient magnitude reflects a lower retention of knowledge within these layers, explaining the insensitivity of knowledge-based tasks to layer removal. Conversely, reasoning processes are closely tied to language itself, meaning the removal of these layers has a more significant impact on such tasks.

\section{Explanation of Industry-Specific Jargons}
We provide detailed explanations for potentially confusing industry-specific jargon mentioned in the paper, ensuring clarity without compromising technical accuracy.

\textbf{The Leaderboard}: A standardized platform (e.g., Hugging Face’s Open LLM Leaderboard) for comparing model performance across benchmarks.

\textbf{MoE (Mixture of Experts)}: A neural network architecture that dynamically routes inputs to a subset of specialized expert models, improving computational efficiency and scalability in large language models (LLMs).

\textbf{VRAM (Video Random Access Memory)}: The GPU's dedicated memory, critical for deploying large language models (LLMs) because its capacity constrains the maximum size of models that can be loaded and run.

\textbf{IQR (Interquartile Range)}: A statistical measure of data spread between the 25th and 75th percentiles, reducing the influence of outliers. Applied in Figure~\ref{fig:performance_size_train} to capture performance fluctuations across model sizes.

\textbf{LLaMA-Factory}: An open-source framework designed for fine-tuning, training, and deploying large language models.

\textbf{LoRA (Low-Rank Adaptation)}: A parameter-efficient fine-tuning technique that uses low-rank matrix decomposition.

\textbf{Impurity (MSE for regression trees)}: A measure of node heterogeneity in decision trees used to determine feature splits. 
In regression, impurity is measured by mean squared error (MSE), and feature importance comes from its weighted decrease after splitting. 
(For classification, impurity is usually measured by Gini impurity or entropy.)

\end{document}